\documentclass[runningheads]{llncs}

\usepackage{eccv}
\usepackage{eccvabbrv}
\usepackage{graphicx}
\usepackage{booktabs}
\usepackage[accsupp]{axessibility}
\usepackage{wrapfig}
\usepackage{caption}
\usepackage{subcaption}
\usepackage{multirow}
\usepackage{array}
\usepackage[table, dvipsnames]{xcolor}
\usepackage{placeins}
\usepackage{pifont}
\definecolor{lightgray}{rgb}{0.86, 0.86, 0.86}
\usepackage{siunitx}
\usepackage{hyperref}
\usepackage[export]{adjustbox}
\usepackage{orcidlink}

\begin{document}

\title{Cast and Attached Shadow Detection via Iterative Light and Geometry Reasoning}
\titlerunning{Attached Shadow Detection via Iterative Reasoning}

\author{
Shilin Hu\inst{1}\orcidlink{0009-0006-6002-1403}\and
Jingyi Xu\inst{1}\and
Sagnik Das\inst{1}\and
Dimitris Samaras\inst{1}\textsuperscript{$\dagger$}\orcidlink{0000-0002-1373-0294}\and
Hieu Le\inst{2}\textsuperscript{$\dagger$}\orcidlink{0000-0001-7855-2778}
}

\authorrunning{S.~Hu et al.}

\institute{
Stony Brook University, Stony Brook NY 11794, USA \and UNC Charlotte, Charlotte NC 28223, USA\\
\email{\{shilhu,jingyixu,sadas,samaras\}@cs.stonybrook.edu},
\email{hle40@charlotte.edu}
}

\maketitle
\begingroup
\renewcommand{\thefootnote}{}
\NoHyper
\footnotetext{\textsuperscript{$\dagger$} Equal advising.}
\endNoHyper
\endgroup

\begin{abstract}

Shadows encode rich information about scene geometry and illumination, yet existing methods either predict a unified shadow mask or overlook attached shadows entirely. 
We address this gap by proposing a framework for jointly detecting cast and attached shadows through explicit physical modeling of light direction and surface geometry under a dominant directional-light setting. 
Our approach is grounded in a simple observation: surfaces facing away from the light source tend to fall into shadow. We exploit the reciprocal relationship between shadow formation and light estimation to construct a closed feedback loop, a dual-module architecture in which a shadow detection module and a light estimation module iteratively refine each other. At each pass, updated light estimates, together with surface normals, produce partial attached shadow maps that guide detection, while improved shadow predictions sharpen light estimation. To support training and evaluation, we introduce a dataset of 1,458 images with manually annotated cast and attached shadow masks sourced from three existing benchmarks. Experiments demonstrate that our proposed method outperforms prior methods, with at least a 33\% reduction in attached-shadow BER, while maintaining strong full-shadow and cast-shadow performance.
Project page: \url{https://shilin21.github.io/attached_detection/}

\end{abstract}
    
\section{Introduction}
\label{sec:intro}
\newcommand{\cmark}{\ding{51}}  
\newcommand{\xmark}{\ding{55}}  

\begin{figure}[t]
  \centering

  \begin{adjustbox}{valign=t,minipage=0.44\linewidth}
    \centering
    \captionof{table}{\textbf{Comparison of recent shadow detection datasets.}
    Most existing datasets omit attached shadows or annotate them only implicitly,
    while our dataset provides explicit cast/attached shadow annotations as separate classes.}
    \label{tab: datasets}

    \small
    \setlength{\tabcolsep}{3pt}
    \begin{tabular}{lccc}
      \toprule
      \textbf{Dataset} & \textbf{Att.} & \textbf{\#Imgs} & \textbf{Split} \\
      \midrule
      SBU~\cite{Vicente-etal-ECCV16}           & \cmark\,(occ.) & 4.1K  & \xmark \\
      ISTD~\cite{Wang_2018_CVPR}               & \xmark          & 1.9K  & \xmark \\
      SOBA~\cite{wang2020instance}             & \xmark          & 1.0K  & \xmark \\
      ViSha~\cite{Chen2021TriplecooperativeVS} & \xmark          & 11.7K & \xmark \\
      SILT~\cite{yang2023silt}                 & \cmark          & 0.6K  & \xmark \\
      CUHK~\cite{Hu2019RevisitingSD}           & \cmark          & 10.5K & \xmark \\
      \textbf{Ours}                            & \cmark          & 1.5K  & \cmark \\
      \bottomrule
    \end{tabular}
  \end{adjustbox}
  \hfill
  \begin{adjustbox}{valign=t,minipage=0.54\linewidth}
    \centering
    \includegraphics[width=\linewidth,height=5.2cm,keepaspectratio]{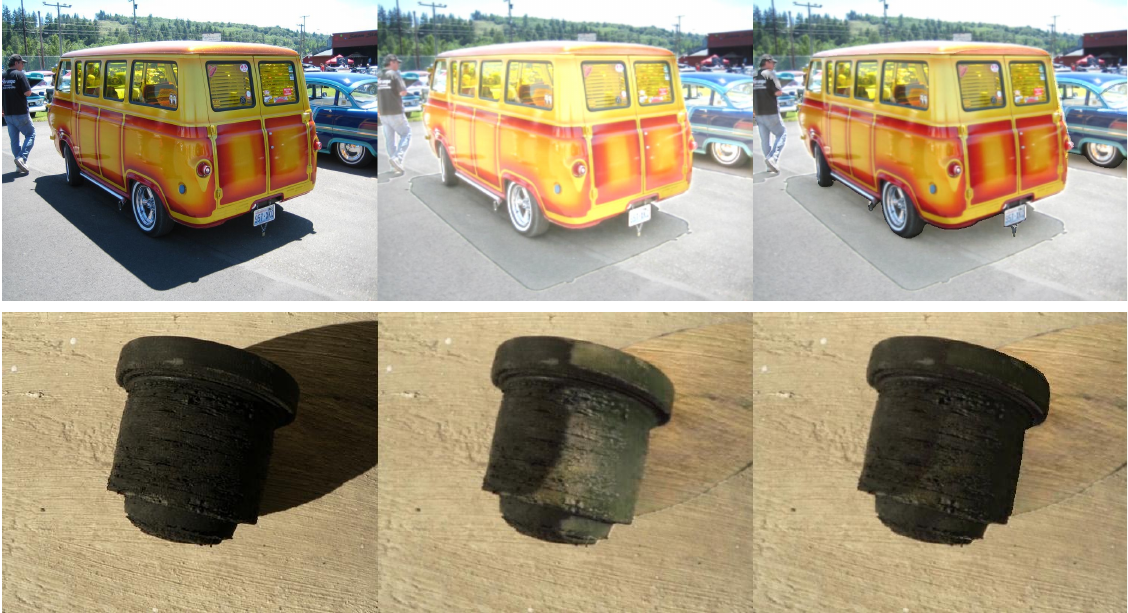}

    {\small
      \makebox[0.32\linewidth]{Input}%
      \makebox[0.35\linewidth]{Unified Mask}%
      \makebox[0.32\linewidth]{Separate Masks}%
    }

    \captionof{figure}{\textbf{Effect of shadow mask granularity on removal quality.}
    Using a unified shadow mask, ShadowFormer~\cite{guo2023shadowformer} over-attenuates object regions.
    Using separate cast and attached shadow masks sequentially enables targeted correction and recovers natural object color.}
    \label{fig: motivation}
  \end{adjustbox}

\end{figure}
\setcounter{footnote}{0}

Shadows define how we perceive shape, depth, and illumination in visual scenes.
They are commonly categorized into two types: cast shadows, which appear on external surfaces and reveal spatial relations between objects \cite{Lalonde10, panagopoulos2009robust, panagopoulos2012simultaneous, Sunkavalli2010VisibilitySU}, and attached shadows, which form directly on an object's surface when light is blocked due to the surface's geometry and orientation relative to the light source \cite{Zhu2012ObjectbasedCA, Shadow_tracking}. Together, these two shadow types encode complementary cues about scene structure and illumination, enriching visual realism and 3D perception \cite{Chuang2003, Wang2020PeopleAS, Zhang2019ImprovingSS, yang2025mli}.

However, most modern shadow detection methods predict a unified shadow mask~\cite{Vicente-etal-ECCV16,yang2023silt,Hu2019RevisitingSD}, or ignore attached shadows entirely~\cite{Chen2021TriplecooperativeVS,Wang_2018_CVPR,wang2020instance} (see \cref{tab: datasets}). As a result, \textbf{explicitly} modeling attached shadows remains largely unexplored in the literature, despite their potential usefulness in downstream applications. In 3D shape reconstruction, attached shadows provide strong cues about local surface orientation and therefore reveal valuable information about the underlying geometry of objects. 
In shadow removal, separate cast and attached masks enable more targeted correction of receiving surfaces and object regions. As a simple illustration, applying an existing method~\cite{guo2023shadowformer} with separate cast and attached masks improves removal quality over using a single unified mask (see \cref{fig: motivation}).

\begin{figure}[!t]
  \centering

  \begin{subfigure}[t]{0.499\linewidth}
    \centering
    \includegraphics[width=\linewidth]{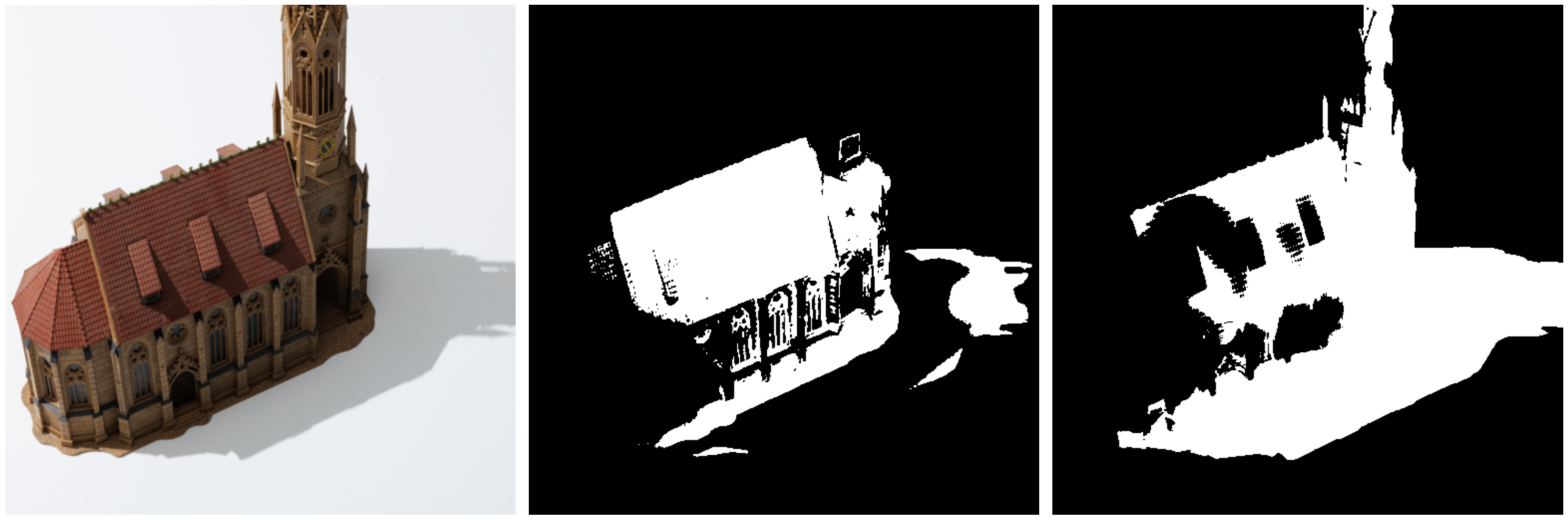}
    {\small
      \makebox[0.33\linewidth]{(a)Input}%
      \makebox[0.33\linewidth]{(b)MTMT\cite{chen2020multi}}%
      \makebox[0.33\linewidth]{(c)SDCM\cite{zhu2022single}}%
    }
  \end{subfigure}\hfill
  \begin{subfigure}[t]{0.499\linewidth}
    \centering
    \includegraphics[width=\linewidth]{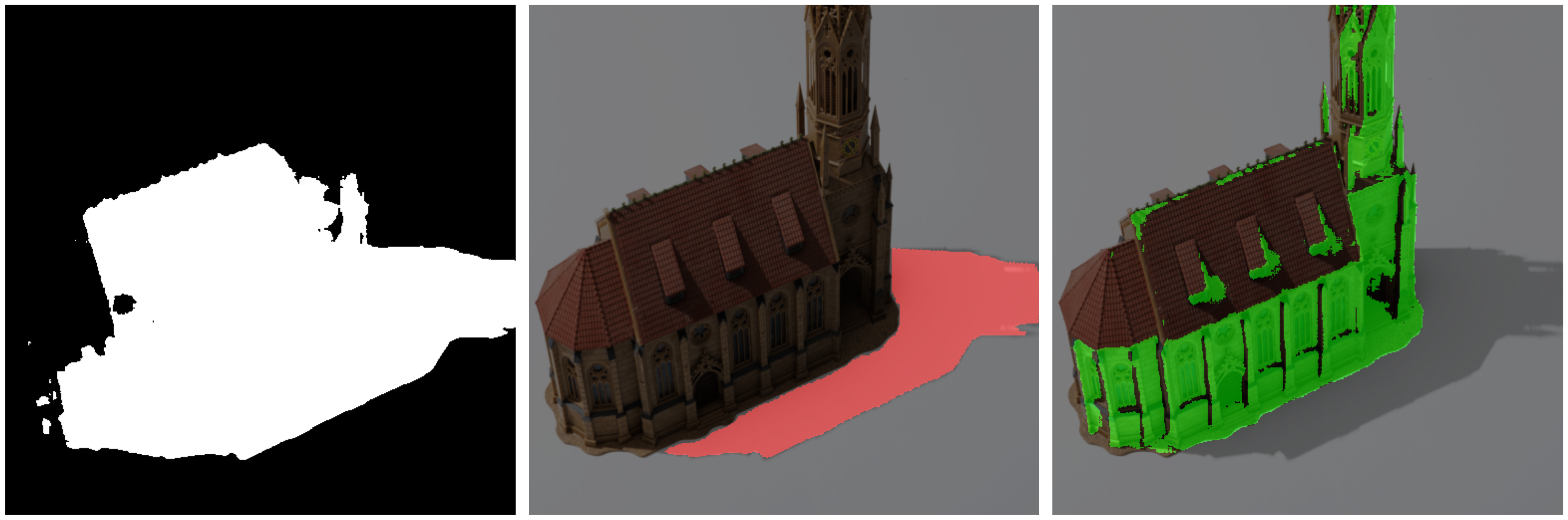}
    {\small
      \makebox[0.33\linewidth]{(d)SILT\cite{yang2023silt}}%
      \makebox[0.33\linewidth]{(e)Ours Cast}%
      \makebox[0.33\linewidth]{(f)Ours Att.}%
    }
  \end{subfigure}
  \caption{Existing shadow detection methods primarily focus on cast shadow detection and do not differentiate between cast and attached shadow types (b, c, d). In contrast, our method accurately segments cast and attached shadows separately (e, f).}
  \label{fig: teaser}
\end{figure}

Yet attached shadow detection is far more than a simple extension of existing shadow detection approaches. As we will show experimentally, naive adaptations of current methods \cite{zhu18b, Hu2019RevisitingSD, chen2020multi, zhu2021mitigating, zhu2022single, yang2023silt} fall short; see \cref{fig: teaser} for typical predictions. 
This difficulty arises because cast and attached shadows are governed by different formation mechanisms: cast shadows are often driven by object boundaries and produce relatively contiguous regions of contrast, amenable to appearance-based detection. Attached shadows, by contrast, arise from the interplay between surface orientation and illumination. They follow object geometry closely, often appearing fragmented with weak contrast, and resist detection without explicit reasoning about scene geometry and lighting.

We therefore introduce a framework that jointly detects cast and attached shadows by modeling their physical dependence on light direction and surface geometry. The key insight is simple: under direct illumination, surfaces facing away from the light receive no direct light and therefore form attached shadows. This relationship works in both directions.
Given a light direction\footnote{We assume a single dominant directional light source.} and surface normals, we can estimate a partial attached-shadow map. Conversely, given observed shadows and scene geometry, we can partially reason about the light direction. We exploit this reciprocal relationship through a closed-loop, dual-module architecture, in which a shadow detection module and a light estimation module mutually refine each other across iterative passes.

To support training and evaluation, we contribute a dataset of 1,458 images with manually annotated cast and attached shadow masks, drawn from three existing benchmarks \cite{vasluianu2024ntire,wang2020instance,Hu2019RevisitingSD}. Experiments confirm that our iterative, geometry- and light-aware formulation outperforms prior methods on both shadow types.

Our contributions are:
\begin{itemize}
  \item A joint framework for detecting cast and attached shadows through shared learning with light estimation.
  \item An iterative refinement scheme in which lighting and shadow cues mutually reinforce each other.
  \item A curated dataset with separate annotations for cast and attached shadows, enabling standardized evaluation.
  \item Our method outperforms previous methods on both shadow types, with at least a 33\% reduction in attached-shadow BER.
\end{itemize}
    
\section{Related Work}
\label{sec:related}
Shadow detection and removal \cite{Qu_2017_CVPR, hu_iccv2019mask, Cun2020TowardsGS, le2021physics, guo2023shadowdiffusion, li2023leveraging, jin2024des3, Le_2020_ECCV, Le-etal-ICCV19, Hu2024ShadowRR} are fundamental tasks in computer vision, where accurate shadow estimation is crucial for tasks such as scene understanding~\cite{Le2016GeodesicDH, LeICCV2017, Le_2019_CVPR_Workshops, Le_RS22} and generation \cite{Xu2024ASQ, Xu2023FSOD, Xu2023ZeroShotOC, Xu2022GeneratingRS, Wu2025ImportanceBasedTM, JingyiICCV21}.
Earlier shadow detection studies \cite{sato2003illumination, okabe2009attached, panagopoulos2009robust, panagopoulos2012estimating, panagopoulos2011cvpr, panagopoulos2011iccv} extensively explored the relationship between lighting and shadows. Panagopoulos \etal \cite{panagopoulos2012simultaneous} were among the first to jointly model shadow detection and light estimation using coarse scene geometry. However, their approach relied on simplified geometry and idealized physical models, which limited its ability to handle the complexities and variations in real-world images.

Recent deep learning advancements have significantly improved shadow detection \cite{Wang_2018_CVPR, zhu18b, Hu_2018_CVPR, m_Le-etal-ECCV18, Zheng2019DistractionAwareSD, Hu2019RevisitingSD, chen2023sam}.
Ding \etal \cite{Ding2019ARGANAR} introduced an attentive recurrent GAN that iteratively detects and removes shadows. Chen \etal \cite{chen2020multi} leveraged multiple cues from shadow regions, edges, and shadow counts, using a student-teacher model. Zhu \etal \cite{zhu2021mitigating} noted that deep shadow detectors were overly reliant on intensity cues and proposed a feature decomposition and reweighting approach to reduce intensity bias. SDCM \cite{zhu2022single} used a complementary mechanism to jointly detect shadow and non-shadow regions by transferring deactivated intermediate features between branches. Sun \etal \cite{sun2023adaptive} focused on preserving multiscale contrast information by mapping RAW images to sRGB representations of varying intensities. More recently, SwinShadow \cite{wang2024swinshadow} leveraged a transformer-based architecture with a shifted window mechanism to better detect adjacent shadows by capturing hierarchical contextual details.
In addition, several approaches have targeted more specialized tasks. \cite{wang2020instance, wang2021single, wang2022instance} focused on the instance shadow detection task, concentrating on cast shadows associated with annotated objects. 
Wang \etal \cite{wang2025metashadow} proposed MetaShadow, which studies object-centered shadow reasoning by jointly addressing shadow detection, removal, and synthesis within a unified framework.

While these recent methods have significantly advanced shadow detection, they do not explicitly address the unique challenge of detecting attached shadows.
Several methods and datasets acknowledge the importance of attached shadows. Yang \etal \cite{yang2023silt} annotated attached shadows for the SBU-test set and proposed a method that uses predicted shadows as updated training labels. Hu \etal \cite{Hu2019RevisitingSD} introduced a benchmark in which a large portion of the shadows are attached shadows. However, these approaches still treat cast and attached shadows uniformly, ignoring their distinctions.

Estimated lighting has been widely used in computer vision tasks. Existing illumination representations can be broadly categorized into three types: environment maps \cite{somanath2021hdr, verbin2024eclipse}, light probes \cite{bai2023local, phongthawee2024diffusionlight}, and parameterized lighting models \cite{gardner2019deep, zhang2019all, dastjerdi2023everlight}. 
In this paper, we adopt a three-dimensional light-direction parameterization to model a dominant distant directional light, typical of outdoor daylight, and to provide a controlled, interpretable link between illumination and scene geometry. Importantly, shadow understanding under this single-direction assumption remains actively studied \cite{xing2024video}, and attached shadow detection is still challenging even in this setting.

\section{Method}
\subsection{Framework Overview}
Our framework detects both cast and attached shadows by jointly learning shadow detection and light estimation from a single RGB image and its corresponding normal map (\cref{fig: architecture}).
We define attached shadows as shadows on the object itself (typically on surface regions oriented away from the light), and cast shadows as shadows projected onto external surfaces.
Attached shadows depend on surface orientation and lighting, motivating our geometry- and light-aware design.
We assume a single scene-wide directional light shared across the image, and define $\boldsymbol{\ell}$ to point from the light toward the scene.

Our shadow detection module is formulated as a multi-class segmentation task over $\{\text{bg}, \text{cast}, \text{attached}\}$, taking the RGB image and normal map as input.
Our light estimation module leverages the normal map and predicted shadows to estimate the light direction as a three-dimensional vector.
Using the estimated light direction and surface normals, we generate a geometry-derived \emph{partial attached shadow map} from a simple normal-light alignment cue.
This partial map is incorporated as an additional input to refine shadow detection, while the predicted shadows provide auxiliary cues for light estimation.
The two modules operate in a closed loop and are iteratively refined over multiple forward passes, improving both shadow detection and light estimation.

\begin{figure}[!t]
    \centering
    \includegraphics[width=\linewidth]{pics_png/overview.pdf}
    \caption{\textbf{Framework overview.} Our network consists of a shadow detection module and a light estimation module. The detector predicts cast and attached shadows, while the light module estimates a 3D light direction and generates a geometry-based partial attached shadow map. This partial map is fed back to the detector as an additional input during iterative training, enabling progressive refinement.}
    \label{fig: architecture}
\end{figure}

\subsection{Iterative Training}
Shadow and illumination serve as mutual cues for estimating each other~\cite{panagopoulos2012simultaneous}. Similarly, our framework jointly and iteratively refines shadow detection and light estimation.

Each training step consists of multiple iterations. In each iteration, we use the predicted light direction and surface normals to generate a \emph{partial attached shadow map} derived from geometry and light direction.
This map highlights pixels that are likely to belong to attached shadows based purely on local surface orientation and the light direction.
It is not a complete attached shadow mask because it does not model visibility or geometric blocking; thus, it cannot capture cases where one surface region blocks light from reaching another.
Still, it provides reliable local cues near contact regions and boundaries, and is fed to the shadow detection module in the next iteration as an additional input to predict the complete cast and attached shadow masks.
Because this map is generated from the current light estimate during training, the detector learns to use imperfect geometry cues.

Specifically, given a light direction vector $\boldsymbol{\ell} \in \mathbb{R}^3$ pointing from the light source toward the surface and surface normals $\boldsymbol{n} \in \mathbb{R}^{H\times W\times 3}$, the partial attached shadow map, denoted as $\mathcal{M}_{\text{attached}}$, includes surface points oriented away from the light source as:
\begin{equation}
    \mathcal{M}_{\text{attached}}=
    \begin{cases}
    1, & \boldsymbol{n}\cdot\boldsymbol{\ell} > 0 \\
    0, & \boldsymbol{n}\cdot\boldsymbol{\ell} \le 0
    \end{cases}
    \label{eq: attached}
\end{equation}

This geometry-based map serves as a coarse attached-shadow prior.
As the estimated light direction becomes more accurate across iterations, the map is progressively refined, leading to improved shadow detection and, in turn, better light estimation.
In the first iteration, the partial attached-shadow map is set to an all-ones map.

\subsection{Training Objectives}
We jointly optimize the two modules with complementary objectives for shadow detection and light estimation.
The shadow loss enforces accurate and well-separated segmentation of cast and attached shadows, while the light loss supervises the estimated light direction and regularizes it for physical plausibility.
Together, these objectives encourage stable convergence of the iterative refinement process described above.

\paragraph{Shadow Detection Loss.}
We supervise the shadow detector at two complementary levels.
First, for global shadow segmentation (shadow vs.\ background), we aggregate subclass logits with the log-sum-exp (LSE): $\mathrm{LSE}(a,b) = \log(e^{a}+e^{b})$.
Let $z_{\mathrm{bg}}, z_{\mathrm{cast}}, z_{\mathrm{att}}\in\mathbb{R}$ be per-pixel logits; we define:
\begin{equation}
  s = \mathrm{LSE}(z_{\mathrm{cast}},\, z_{\mathrm{att}}) - z_{\mathrm{bg}}.
\end{equation}
We match $s$ to the union ground-truth shadow mask $y_{\mathrm{union}}\in\{0,1\}$ using a binary cross-entropy (BCE) term plus a Dice \cite{sudre2017generalised} regularizer:
\begin{equation}
  \mathcal{L}_{\mathrm{seg}}
  =
  \mathrm{BCE}(s,\, y_{\mathrm{union}})
  +
  \lambda_{\mathrm{Dice}}\,\mathrm{Dice}(s,\, y_{\mathrm{union}}).
\end{equation}

Second, to distinguish shadow types (cast vs.\ attached), we apply a standard multi-class cross-entropy (CE) on the full logits with per-pixel labels $y_{\mathrm{type}}\in\{\text{bg},\text{cast},\text{att}\}$.
In addition, we impose a class-conditional margin on the cast vs.\ attached logits to sharpen their separation.
Let $d^{(i)} = z^{(i)}_{\text{cast}} - z^{(i)}_{\text{att}}$, and let $\Omega_{\text{cast}}$ and $\Omega_{\text{att}}$ be the index sets of pixels labeled cast and attached, respectively.
With margin $m=0.2$, we have:
\begin{equation}
\mathcal{L}_{\text{dist}}
=\frac{1}{|\Omega_{\text{cast}}|}
\sum_{i \in \Omega_{\text{cast}}} \max\!\big(0,\, m - d^{(i)}\big)
+ \frac{1}{|\Omega_{\text{att}}|}
\sum_{i \in \Omega_{\text{att}}} \max\!\big(0,\, m + d^{(i)}\big),
\end{equation}
\begin{equation}
\mathcal{L}_{\text{type}}
=
\mathrm{CE}(\mathbf{z},\, y_{\mathrm{type}})
+
\lambda_{\text{dist}}\,\mathcal{L}_{\text{dist}}.
\end{equation}
Together, this supervision encourages robust recovery of overall shadow extent while explicitly teaching the model to separate attached from cast shadows.
The total shadow loss is therefore:
\begin{equation}
\mathcal{L}_{\text{shadow}}
=
\mathcal{L}_{\text{seg}}
+
\mathcal{L}_{\text{type}}.
\end{equation}

\paragraph{Light Estimation Loss.}
We constrain the light estimator with three complementary losses.
First, we encourage light cues to localize attached shadows by aligning $\mathcal{M}_{\text{attached}}$ with the ground-truth attached shadow mask $y_{\text{att}}\in\{0,1\}$ via a weighted BCE.
Second, we regress the predicted light direction $\hat{\boldsymbol{\ell}}$ to a heuristic target $\boldsymbol{\ell}^\star$ (from geometry; see \cref{sec: dataset} for details) with an $\ell_1$ loss.
Third, we enforce a unit norm:
\begin{equation}
\mathcal{L}_{\text{att}}=\mathrm{BCE}_w(\mathcal{M}_{\text{attached}},\,y_{\text{att}}),\quad
\mathcal{L}_{\text{dir}}=\big\|\hat{\boldsymbol{\ell}}-\boldsymbol{\ell}^\star\big\|_{1}
,\quad
\mathcal{L}_{\text{unit}}=(\|\hat{\boldsymbol{\ell}}\|_2-1)^2.
\end{equation}
The overall light loss is then:
\begin{equation}
    \mathcal{L}_{\text{light}}
    = \lambda_{\text{att}}\mathcal{L}_{\text{att}}
    + \lambda_{\text{dir}}\mathcal{L}_{\text{dir}}
    + \lambda_{\text{unit}}\mathcal{L}_{\text{unit}}.
\end{equation}
Finally, the full training objective is:
\begin{equation}
\mathcal{L}_{\text{total}}
=
\mathcal{L}_{\text{shadow}}
+
\mathcal{L}_{\text{light}}.
\end{equation}

\newlength{\headw}   
\newlength{\cwA}     

\begin{figure}[t]
\centering
\begin{minipage}{0.8\linewidth}
  \setlength{\headw}{0.04\linewidth}
  \includegraphics[width=\linewidth]{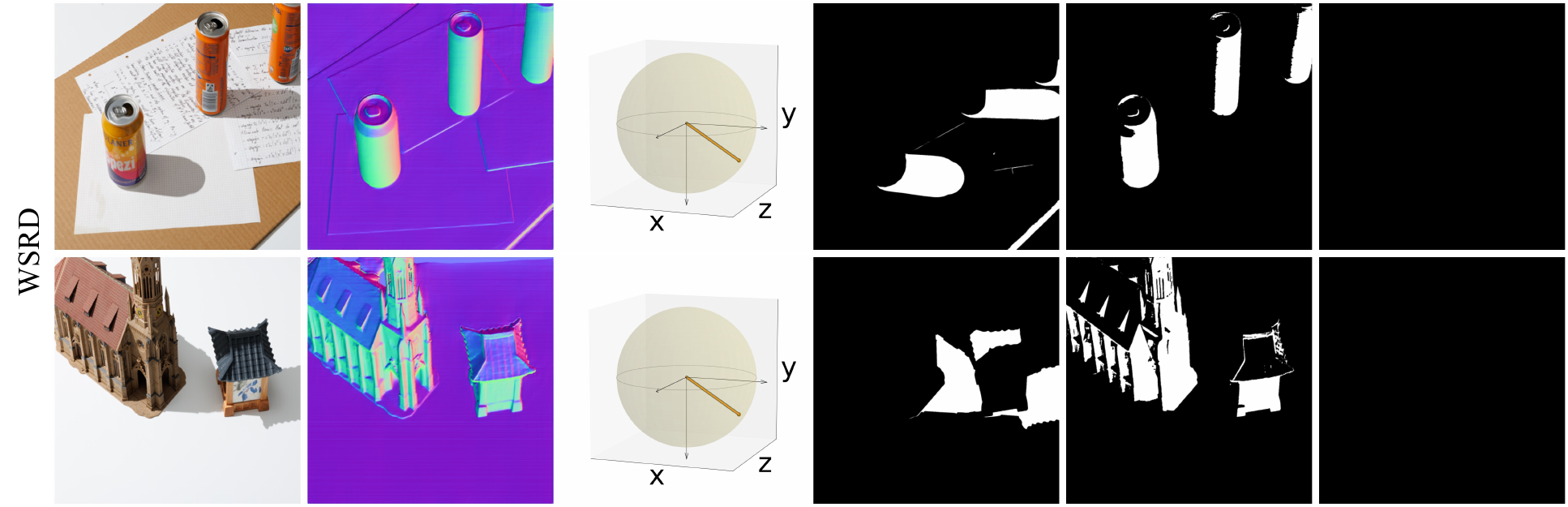}
  \includegraphics[width=\linewidth]{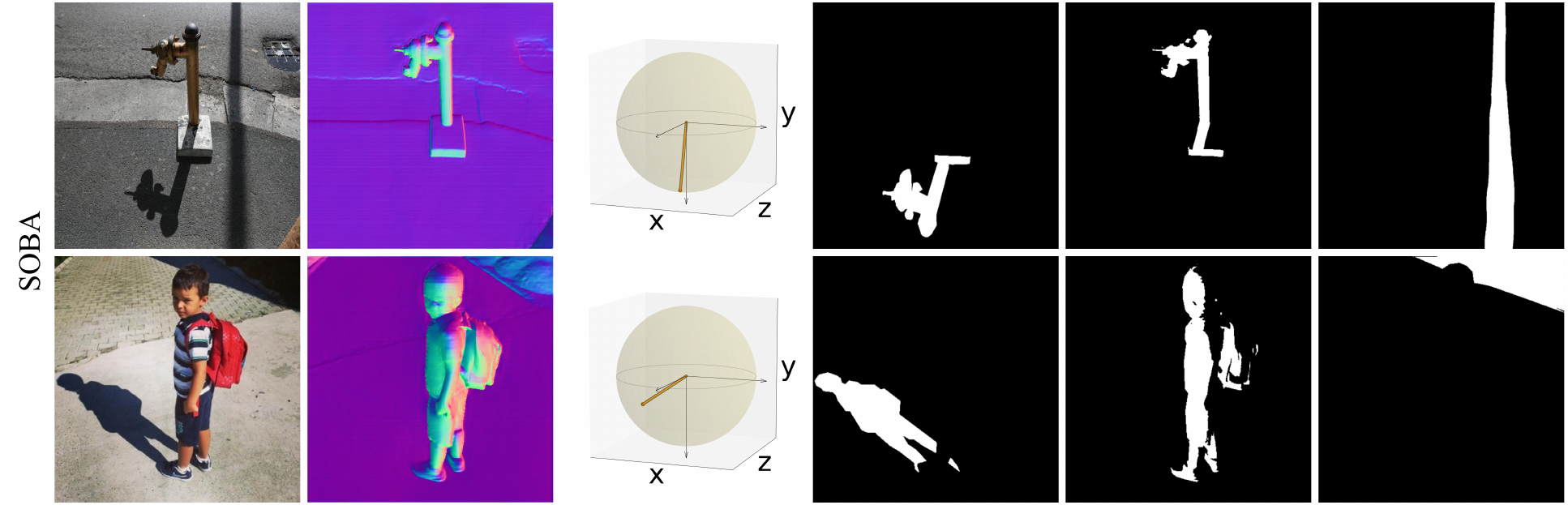}
  \includegraphics[width=\linewidth]{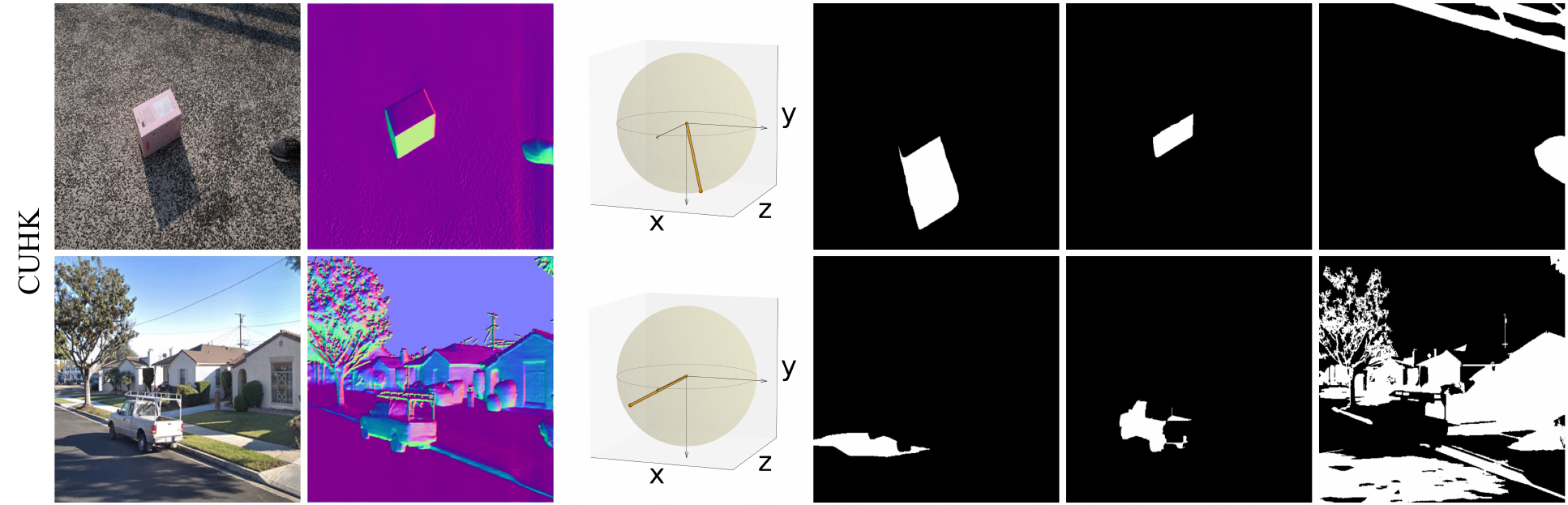}
  \setlength{\cwA}{\dimexpr(\linewidth-\headw)/6\relax}
{\small
  \makebox[\headw][c]{\ }%
  \makebox[\cwA][c]{Image}%
  \makebox[\cwA][c]{Normal}%
  \makebox[\cwA][c]{Light}%
  \makebox[\cwA][c]{Cast}%
  \makebox[\cwA][c]{Attached}%
  \makebox[\cwA][c]{Undefined}%
}
\end{minipage}
\caption{\textbf{Examples of our proposed dataset.} We curate the dataset from WSRD~\cite{vasluianu2024ntire}, SOBA~\cite{wang2020instance}, and CUHK~\cite{Hu2019RevisitingSD}. Each image is paired with its normal map, light direction, and cast/attached/``undefined'' shadow masks. Light direction is camera-centric: $+x$ right, $+y$ down, $+z$ inward.}
\label{fig: dataset}
\end{figure}

\section{Dataset}
\label{sec: dataset}
We introduce the first dataset specifically curated for cast and attached shadow detection, consisting of 1,166 training and 292 test images.
Each image is paired with (i) a surface normal map, (ii) a ray-based heuristic 3D light direction, (iii) binary cast and attached shadow masks, and (iv) an ``undefined'' shadow mask.
\cref{fig: dataset} shows example annotations.

Our dataset is compiled from three public sources:
220 images from WSRD \cite{vasluianu2024ntire} (paired shadow/shadow-free images),
710 images from SOBA \cite{wang2020instance} (shadow images with instance-level cast masks),
and 528 images from CUHK \cite{Hu2019RevisitingSD} (scene-level full-shadow masks).

Accurately annotating cast and attached shadows per pixel in natural scenes is highly labor-intensive due to mixed shadow types.
To simplify annotation, we restrict fine-grained labeling to foreground objects: we manually mark the cast shadows and attached shadows associated with the annotated objects.
All remaining shadow pixels not attributable to these objects are labeled as \textit{undefined shadows}.
We use undefined shadows only for full-shadow supervision and evaluation, and exclude them from cast/attached training and evaluation.
This avoids ambiguous ownership and provides a fairer assessment of cast/attached separation.
Additional details on data preparation are provided in the supplementary material.

\paragraph{Normal Maps.}
Since geometry is crucial for identifying attached shadows, we use a recent foundation model \cite{yang2024depth} to predict relative depth, which we convert to normal maps.

\paragraph{Object Masks.}
We extract object masks using BiRefNet \cite{zheng2024bilateral}. These masks are used only for annotation and evaluation.

\paragraph{Cast and Attached Shadow Masks.}
Using the object masks, we manually annotate and refine cast and attached shadow masks for SOBA and CUHK.
For WSRD, which lacks shadow masks, we derive full-shadow masks by color-space subtraction between shadow and shadow-free images, and then annotate cast and attached components.

\paragraph{Light Directions.}
WSRD images are captured under a calibrated fixed-lighting system \cite{vasluianu2024ntire}; therefore, we manually compute ground-truth 3D light directions.
For SOBA and CUHK, we estimate a heuristic 3D direction in two steps.
\textbf{(i) In-plane direction.} We compute a 2D image-plane direction by connecting the centroid of a foreground object to the centroid of its cast shadow.
Under a directional-light assumption, a cast shadow can be viewed as the object translated along the illumination direction and projected onto the receiving surface; thus, the object-to-shadow displacement provides the illumination ray's image-plane projection.
\textbf{(ii) Depth sign.} We infer the third component from the relative depth between the object region and its cast shadow region.
If the shadow lies on a surface deeper than the object, the illumination has a component pointing into the scene; if it lies shallower, the component points toward the camera.
Combining this signed component with the in-plane vector and normalizing yields a 3D unit direction in our camera-centric frame ($+x$ right, $+y$ down, $+z$ inward).

\begin{table}[!t]
\centering
\caption{Quantitative comparison with state-of-the-art methods on our dataset. We report BER$\downarrow$ and F1$\uparrow$ for full, cast, and attached shadows. Methods fine-tuned on our training set are marked with $\dagger$. Best results are in \textbf{bold}.}
\label{tab: sota}
\setlength{\tabcolsep}{5pt}
\renewcommand{\arraystretch}{1.15}
\resizebox{0.65\linewidth}{!}{%
\begin{tabular}{@{}l
S[table-format=2.2]S[table-format=2.2]
S[table-format=2.2]S[table-format=2.2]
S[table-format=2.2]S[table-format=2.2]@{}}
\toprule
& \multicolumn{2}{c}{\textbf{Full}} 
& \multicolumn{2}{c}{\textbf{Cast}}
& \multicolumn{2}{c}{\textbf{Attached}} \\
\cmidrule(lr){2-3}\cmidrule(lr){4-5}\cmidrule(lr){6-7}
\textbf{Methods}
& {\textbf{BER}$\downarrow$} & {\textbf{F1}$\uparrow$}
& {\textbf{BER}$\downarrow$} & {\textbf{F1}$\uparrow$}
& {\textbf{BER}$\downarrow$} & {\textbf{F1}$\uparrow$} \\
\midrule
BDRAR~\cite{zhu18b}                    & 21.29 & 70.53 &  7.23 & 79.64 & 35.41 & 52.32 \\
DSD~\cite{Zheng2019DistractionAwareSD} & 22.53 & 68.75 &  7.83 & 80.10 & 35.93 & 51.74 \\
FSDNet~\cite{Hu2019RevisitingSD}       & 24.46 & 66.09 & 10.51 & 79.16 & 37.53 & 46.24 \\
MTMT~\cite{chen2020multi}              & 33.10 & 50.00 & 15.16 & 74.66 & 42.02 & 38.10 \\
FDRNet~\cite{zhu2021mitigating}        & 23.69 & 65.69 &  8.95 & 71.65 & 34.82 & 55.15 \\
SDCM~\cite{zhu2022single}              & 24.27 & 66.51 &  7.88 & 83.42 & 36.83 & 49.02 \\
SILT~\cite{yang2023silt}               & 20.20 & 71.51 &  4.64 & 80.39 & 32.70 & 60.23 \\
\midrule
BDRAR$^\dagger$                        &  8.15 & 83.95 &  5.18 & 72.86 & 20.75 & 82.69 \\
FSDNet$^\dagger$                       & 10.17 & 85.10 &  5.82 & 81.94 & 19.49 & 80.57 \\
FDRNet$^\dagger$                       &  8.62 & 83.00 &  4.65 & 73.27 & 23.11 & 81.42 \\
SILT$^\dagger$                         &  6.51 & 82.98 &  {\bfseries 4.52} & 68.31 & 26.57 & 80.72 \\
\midrule
Ours                                   &  {\bfseries 6.50} & {\bfseries 86.61} &  5.04 & {\bfseries 85.46} & {\bfseries 12.92} & {\bfseries 86.49} \\
\bottomrule
\end{tabular}%
}
\end{table}

\section{Experiments and Results}

\paragraph{Implementation Details.}
The proposed framework is implemented using PyTorch \cite{paszke2019pytorch}. The shadow detection backbone follows SILT \cite{yang2023silt}, and the light estimation module uses ConvNeXt-S \cite{liu2022convnet}.
We train with Adam \cite{Adam} (learning rate $5\times10^{-4}$, 20 epochs, batch size 4).
We set $\lambda_{\text{Dice}}$, $\lambda_{\text{dist}}$, $\lambda_{\text{att}}$, $\lambda_{\text{dir}}$, and $\lambda_{\text{unit}}$ to $0.1$, $0.2$, $0.4$, $0.5$, and $0.1$, respectively.
Additional details are provided in the supplementary.

\paragraph{Evaluation Methods and Metrics.}
We compare against seven state-of-the-art methods: BDRAR \cite{zhu18b}, DSD \cite{Zheng2019DistractionAwareSD}, FSDNet \cite{Hu2019RevisitingSD}, MTMT \cite{chen2020multi}, FDRNet \cite{zhu2021mitigating}, SDCM \cite{zhu2022single}, and SILT \cite{yang2023silt}.
We evaluate all baselines with their SBU-pretrained weights and additionally fine-tune BDRAR, FSDNet, FDRNet, and SILT on our training set with full-shadow supervision (marked with $\dagger$ in \cref{tab: sota}).

We report the F1 score,
\(F1 = \frac{2TP}{2TP + FP + FN}\),
and the balanced error rate (BER),
\(\mathrm{BER} = 100 \times \bigl(1 - \tfrac{1}{2}\bigl[\tfrac{TP}{TP+FN} + \tfrac{TN}{TN+FP}\bigr]\bigr)\),
where \(TP\), \(TN\), \(FP\), and \(FN\) denote true positives, true negatives, false positives, and false negatives.

For prior baselines that output a single shadow mask, we split their predictions using the object mask. Cast evaluation excludes the undefined region, and attached evaluation is restricted to the object region.

\newlength{\qcw}
\begin{figure}[!t]
  \centering
  \includegraphics[width=\linewidth]{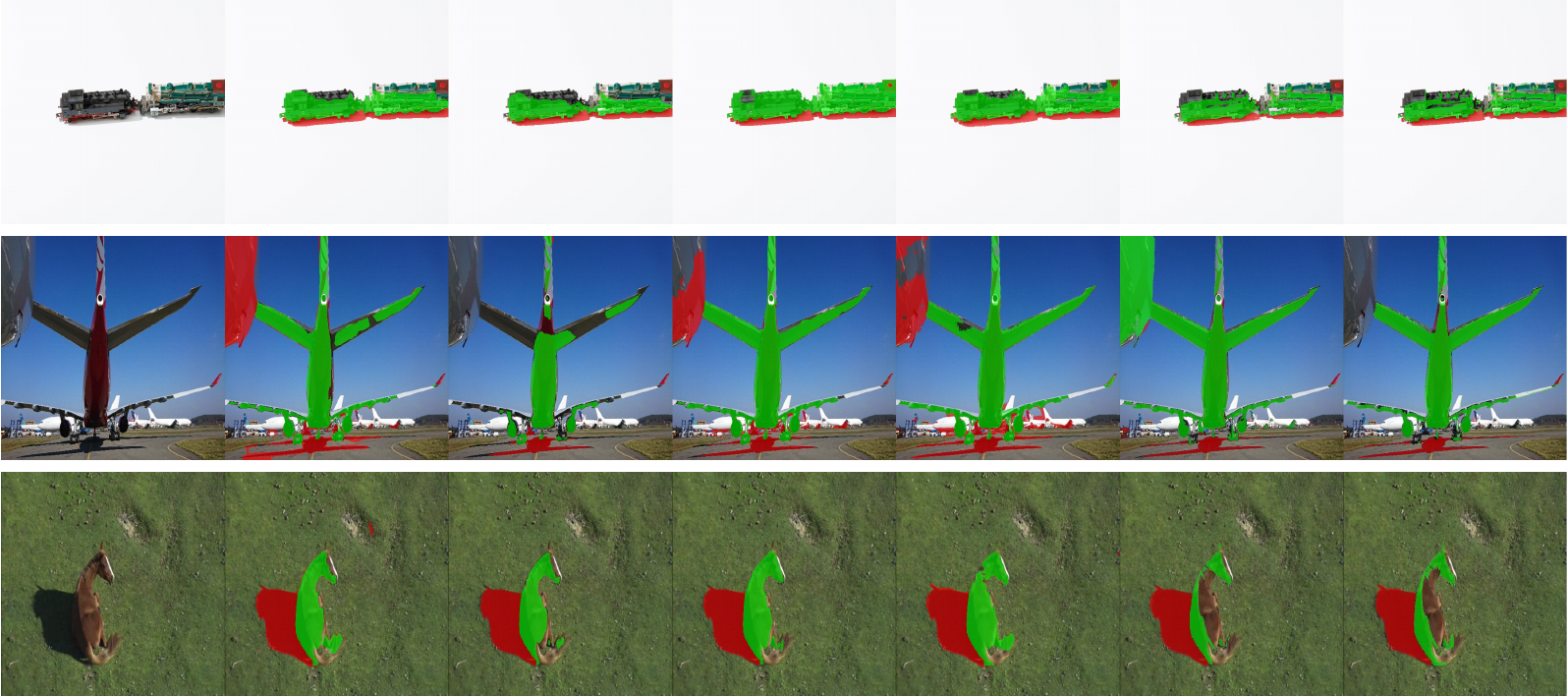}
  \includegraphics[width=\linewidth]{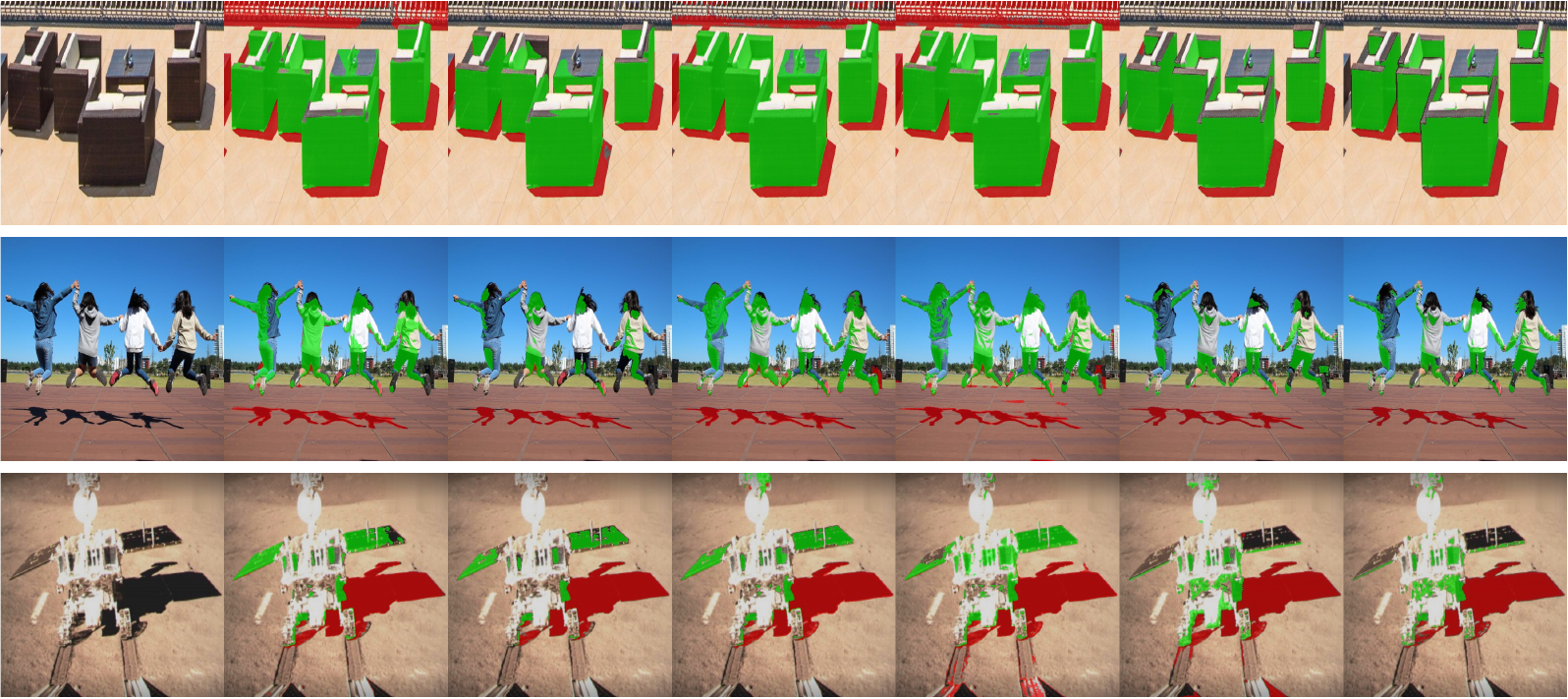}
  \setlength{\qcw}{\dimexpr\linewidth/7\relax}
  \makebox[\qcw][c]{\small Image}%
  \makebox[\qcw][c]{\small BDRAR$^\dagger$}%
  \makebox[\qcw][c]{\small FSDNet$^\dagger$}%
  \makebox[\qcw][c]{\small FDRNet$^\dagger$}%
  \makebox[\qcw][c]{\small SILT$^\dagger$}%
  \makebox[\qcw][c]{\small Ours}%
  \makebox[\qcw][c]{\small GT}%
  \caption{Qualitative comparison of our method with retrained BDRAR~\cite{zhu18b}, FSDNet~\cite{Hu2019RevisitingSD}, FDRNet~\cite{zhu2021mitigating}, and SILT~\cite{yang2023silt} on our dataset. Predicted masks are overlaid on the input image (\textit{green: attached, red: cast}). Our geometry- and light-aware iterative framework produces more accurate attached-shadow boundaries.}
  \label{fig: qual}
\end{figure}

\subsection{Comparison with SOTA methods}
\paragraph{Quantitative Comparisons.}
\cref{tab: sota} reports BER and F1 scores for full, cast, and attached shadows.
Our method achieves the best performance on full and attached shadows in both metrics and the strongest cast F1, while its cast BER remains comparable to prior approaches.

We first evaluate the SBU-pretrained baselines, which are cast-focused.
All methods show a clear gap between cast and attached performance, with attached-shadow BER worse by about 20--30 points.
These models primarily capture cast shadows, which often appear as contiguous, lower-intensity regions, and only detect attached shadows when they visually resemble cast shadows.
This highlights the limitations of cast-biased training data for attached-shadow detection.
We then retrain the baselines on our data using full-shadow supervision.
This narrows the gap, but the baselines still struggle with attached shadows.
A key limitation is that they treat the two shadow types uniformly and ignore that attached shadows are governed by surface orientation and contact rather than occlusion.

In contrast, our method leverages geometry and lighting by using surface normals and estimating a light direction.
These physical cues let the model reason about where attached shadows should occur while preserving cast performance.
As a result, attached shadow BER is reduced by at least 33\% relative to retrained baselines, while maintaining strong performance on full and cast shadows.

\paragraph{Qualitative Comparisons.}
\cref{fig: qual} shows visual comparisons between four retrained baselines and our method.
The baselines produce a single shadow mask, so we use the object mask to split their outputs into cast and attached regions.
Our method follows occluder geometry and the estimated light direction, producing geometry-consistent attached-shadow boundaries on surfaces facing away from the light (rows 1--2).
In contrast, the baselines rely mainly on image appearance and tend to mark dark regions as shadows, even when those surfaces are lit (rows 3--5).
By using surface normals, an estimated light direction, and separate predictions for cast and attached shadows, our method improves attached shadow quality while maintaining strong cast predictions.

\begin{figure}[!t]
  \centering
  \begin{subfigure}[t]{0.495\linewidth}
    \centering
    \includegraphics[width=\linewidth]{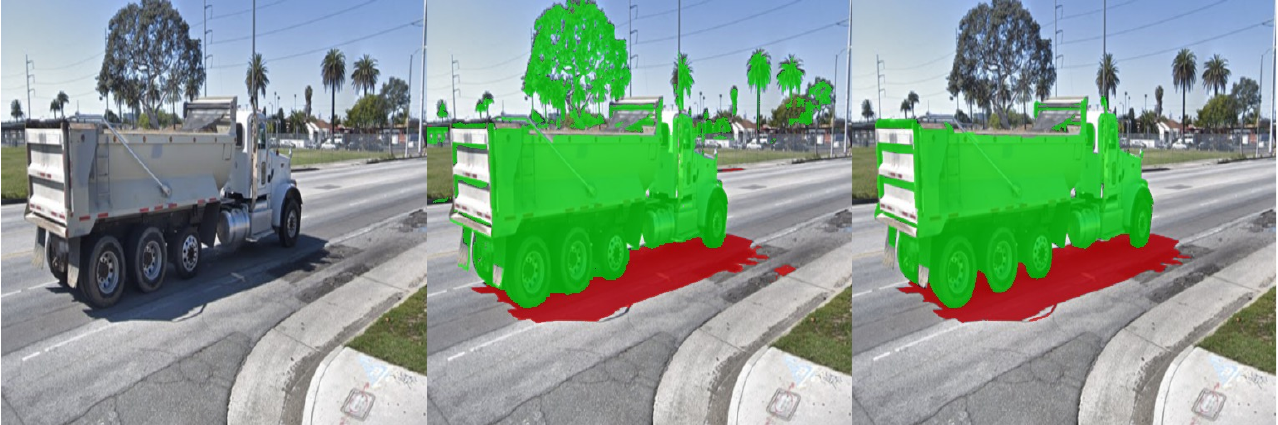}
    {\small
      \makebox[0.33\linewidth]{Image}%
      \makebox[0.33\linewidth]{Our Prediction}%
      \makebox[0.33\linewidth]{GT}%
    }
  \end{subfigure}\hfill
  \begin{subfigure}[t]{0.495\linewidth}
    \centering
    \includegraphics[width=\linewidth]{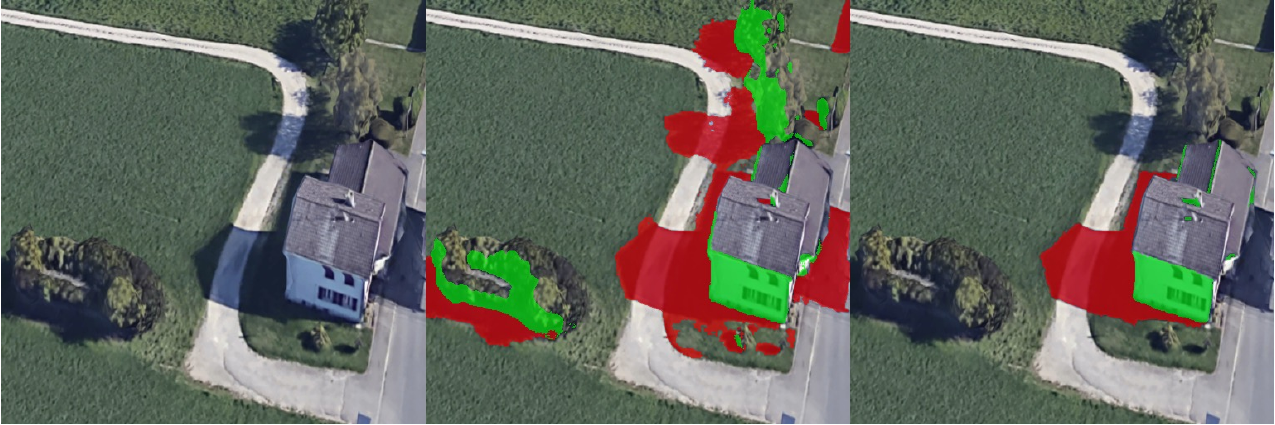}
    {\small
      \makebox[0.33\linewidth]{Image}%
      \makebox[0.33\linewidth]{Our Prediction}%
      \makebox[0.33\linewidth]{GT}%
    }
  \end{subfigure}
  \caption{Examples with foreground-only annotations. Although supervision is limited to the foreground objects, our method produces coherent cast and attached shadow predictions over the full image.}
  \label{fig: overall}
\end{figure}

We further show qualitative results in \cref{fig: overall} where the ground-truth annotations cover only foreground objects, yet our method correctly assigns cast and attached labels across the remaining undefined shadow regions.
This suggests a potential path toward reducing dense annotation effort through pseudo-labels or weak supervision.

\begin{table}[!t]
\centering
\caption{Ablation of our geometry- and light-aware framework. We evaluate the impact of (i) removing surface normals, (ii) varying the number of refinement iterations, and (iii) disabling individual loss terms. We report BER$\downarrow$ and F1$\uparrow$ for full, cast, and attached shadow detection.}
\label{tab: compare}
\setlength{\tabcolsep}{5pt}
\renewcommand{\arraystretch}{1.15}
\resizebox{0.65\linewidth}{!}{%
\begin{tabular}{@{}l
S[table-format=2.2]S[table-format=2.2]
S[table-format=2.2]S[table-format=2.2]
S[table-format=2.2]S[table-format=2.2]@{}}
\toprule
& \multicolumn{2}{c}{\textbf{Full}}
& \multicolumn{2}{c}{\textbf{Cast}}
& \multicolumn{2}{c}{\textbf{Attached}} \\
\cmidrule(lr){2-3}\cmidrule(lr){4-5}\cmidrule(lr){6-7}
\textbf{Setting}
& {\textbf{BER}$\downarrow$} & {\textbf{F1}$\uparrow$}
& {\textbf{BER}$\downarrow$} & {\textbf{F1}$\uparrow$}
& {\textbf{BER}$\downarrow$} & {\textbf{F1}$\uparrow$} \\
\midrule
\textbf{Geometry} \\
No Normal                      & 7.60 & 85.46 & 7.82 & 79.33 & 20.87 & 77.07 \\
\midrule
\textbf{Iterations} \\
Non-iterative                  & 7.57 & 85.99 & 6.14 & 87.75 & 15.10 & 83.91 \\
2 iterations                   & 6.63 & 87.77 & 5.41 & 86.75 & 13.22 & 85.63 \\
\rowcolor{lightgray}3 iterations (final) & 6.50 & 86.61 & 5.04 & 85.46 & 12.92 & 86.49 \\
5 iterations                   & 6.11 & 87.14 & 5.83 & 85.57 & 13.51 & 85.74 \\
\midrule
\textbf{Loss terms} \\
No $\mathcal{L}_{\text{dist}}$ & 7.19 & 86.25 & 6.16 & 85.69 & 13.45 & 85.38 \\
No $\mathcal{L}_{\text{att}}$  & 6.89 & 87.03 & 5.81 & 85.26 & 14.61 & 84.24 \\
No $\mathcal{L}_{\text{dir}}$  & 7.50 & 86.42 & 4.97 & 85.02 & 15.26 & 83.59 \\
\bottomrule
\end{tabular}%
}
\end{table}

\subsection{Ablation Studies}
To validate the effectiveness of our framework design, we conduct ablation studies to analyze the contributions of (i) surface-normal guidance, (ii) the iterative learning scheme, and (iii) each loss component.

\paragraph{Effect of geometry guidance.}
We examine the role of surface normals as geometry guidance.
Removing surface normals reduces our model to a standard multi-class segmentation network trained with class-specific labels.
This substantially degrades performance across metrics, with the largest drop on attached shadows, yielding results comparable to retrained baselines.
This confirms that surface normals provide strong spatial priors for separating cast and attached shadows.

\begin{figure}[!t]
  \centering
  \begin{minipage}{0.8\linewidth}
  \includegraphics[width=\linewidth]{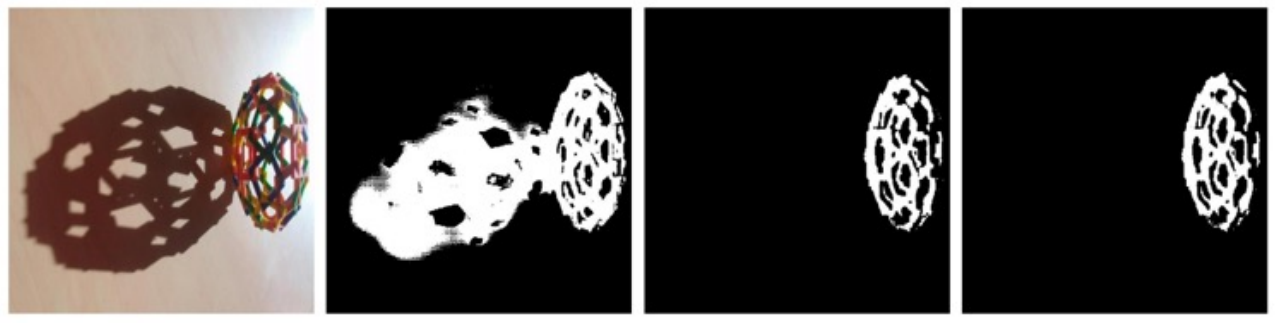}
  \makebox[0.24\linewidth]{\small{\raisebox{1mm}{Image}}}
  \makebox[0.24\linewidth]{\small{\raisebox{1mm}{Att. Iter 1}}}
  \makebox[0.24\linewidth]{\small{\raisebox{1mm}{Att. Iter 2}}}
  \makebox[0.24\linewidth]{\small{\raisebox{1mm}{Att. Iter 3}}}
  \includegraphics[width=\linewidth]{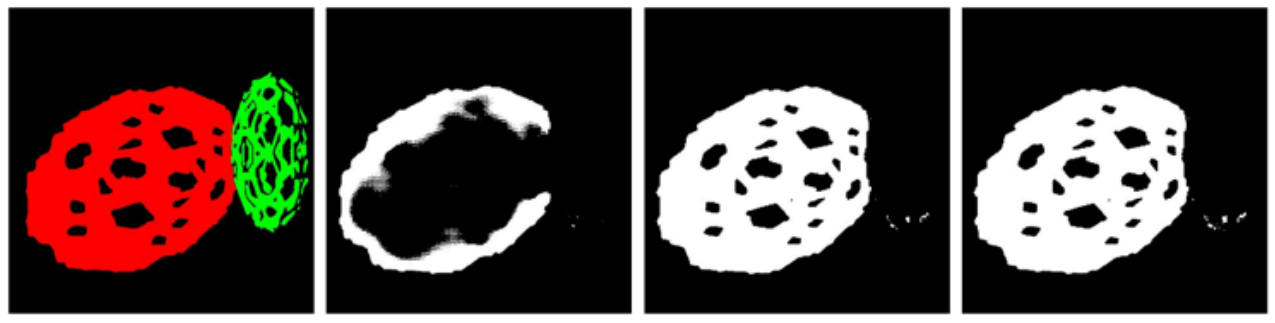}
  \makebox[0.24\linewidth]{\small{\raisebox{1mm}{GT}}}
  \makebox[0.24\linewidth]{\small{\raisebox{1mm}{Cast Iter 1}}}
  \makebox[0.24\linewidth]{\small{\raisebox{1mm}{Cast Iter 2}}}
  \makebox[0.24\linewidth]{\small{\raisebox{1mm}{Cast Iter 3}}}
  \end{minipage}
  \caption{\textbf{Effectiveness of the iterative learning scheme.} Our proposed feedback mechanism progressively refines the shadow predictions, showing clear visual improvements at the second iteration and stabilization at the third.}
  \label{fig: iter}
\end{figure}

\paragraph{Effect of iterative learning.}
We evaluate a non-iterative variant that jointly trains shadow detection and light estimation with normal maps, but does not feed the predicted partial attached map back into the detector.
As shown in \cref{tab: compare}, this setting outperforms the variant without normal guidance, confirming the benefit of geometry awareness.
We then vary the number of refinement iterations (2, 3, and 5).
Iterative feedback improves predictions: two iterations provide the largest gain, and performance saturates at three iterations, as also shown by the intermediate results in \cref{fig: iter}.
Increasing to five iterations slightly improves full-shadow accuracy but reduces class-specific performance, suggesting over-refinement of the partial map.
In terms of efficiency, runtime increases approximately linearly with the number of iterations; our 3-step setting is 3$\times$ slower than the non-iterative model (0.55\,s vs.\ 0.18\,s per image).
Overall, geometry provides the key prior, and iterative refinement further improves segmentation.

\paragraph{Effect of loss design.}
We ablate each loss term individually during training.
Removing any single term reduces performance, indicating that the losses are complementary.
$\mathcal{L}_{\text{dist}}$ encourages separation between cast and attached logits via a class-conditional margin, leading to a clearer decision boundary.
The light-related losses, $\mathcal{L}_{\text{att}}$ and $\mathcal{L}_{\text{dir}}$, guide the light estimator to produce spatially coherent partial maps by aligning outputs with ground-truth cues and enforcing directional consistency.
Together, these objectives promote effective geometry--light interaction and more reliable shadow predictions.

\begin{figure}[!t]
  \centering
  \includegraphics[width=\linewidth]{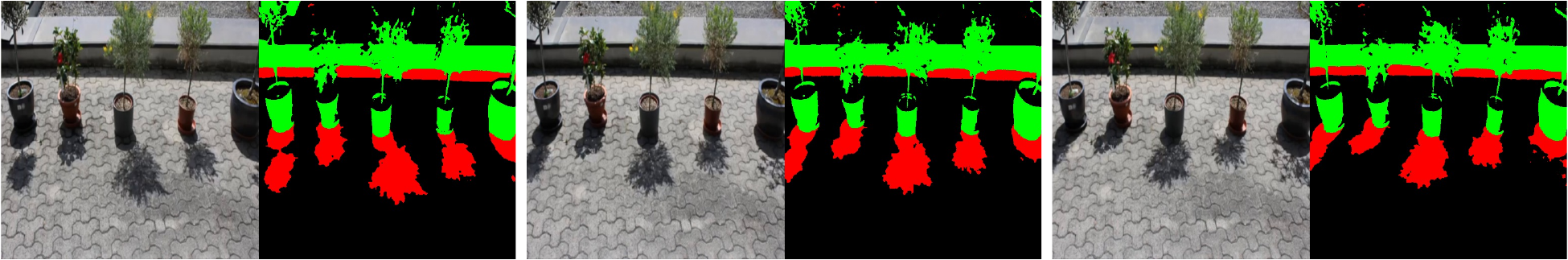}
  {\small
    \makebox[0.33\linewidth]{Frame 120}%
    \makebox[0.33\linewidth]{Frame 200}%
    \makebox[0.33\linewidth]{Frame 280}%
  }
  \caption{\textbf{Cross-dataset generalization on SBU-TimeLapse \cite{le2021physics}.}
  We apply our model to a static-camera video with time-varying shadows and visualize predictions at different time steps.
  The results remain temporally consistent and show coherent cast and attached shadow separation across the sequence.}
  \label{fig:sbu_timelapse}
\end{figure}

\subsection{Cross-dataset Generalization}
To assess generalization beyond our training data, we conduct experiments on SBU-TimeLapse \cite{le2021physics}, a static-scene video dataset with time-varying shadows.
We run our model on each frame independently.
As shown in \cref{fig:sbu_timelapse}, our predictions remain temporally consistent while producing accurate cast and attached shadow separation across the sequence, indicating robustness to temporal shadow variation.
Additional qualitative results are provided in the supplementary.

\begin{figure}[!t]
  \centering
  \includegraphics[width=\linewidth]{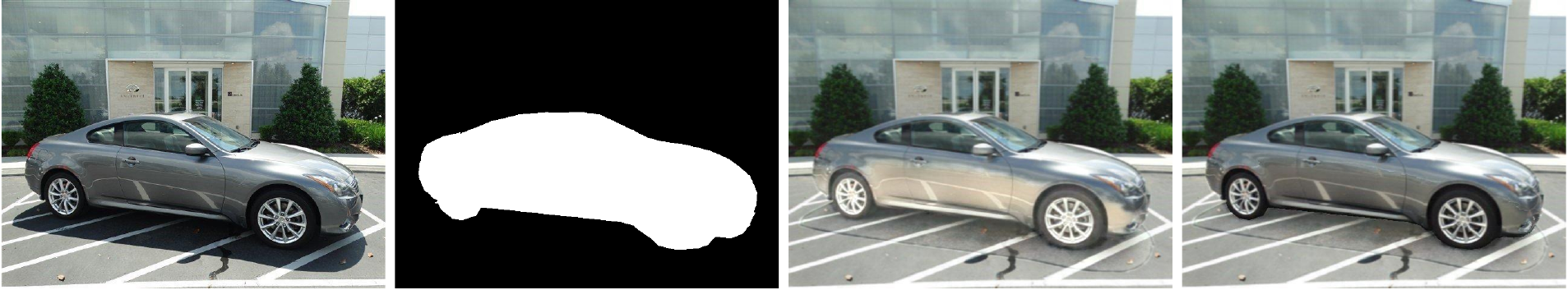}
  {\small
    \makebox[0.24\linewidth]{Image}%
    \makebox[0.24\linewidth]{Object Mask}%
    \makebox[0.24\linewidth]{Result 1}%
    \makebox[0.24\linewidth]{Result 2}%
  }
  \caption{\textbf{User study example.}
  Each question shows the input image and object mask, along with two candidate results:
  one guided by a joint shadow mask and the other guided by separate cast/attached masks with object-region refinement.
  Candidate order is randomized, and participants select the more natural output.}
  \label{fig:user_study}
\end{figure}

\subsection{Shadow Removal Application}
\begin{figure}[!t]
\centering

\begin{adjustbox}{valign=t,minipage=0.38\linewidth}
\centering
\captionof{table}{User study on shadow removal. Participants were asked to choose the more natural result between the joint-mask guidance (Joint) and separate-mask guidance with attached-region refinement (Separate).}
\label{tab: user-study}

\resizebox{0.9\linewidth}{!}{%
  \begin{tabular}{lc}
    \toprule
    \textbf{Method} & \textbf{Preference (\%)} \\
    \midrule
    Joint     & 19.3 \\
    Separate  & 80.7 \\
    \bottomrule
  \end{tabular}%
}

\end{adjustbox}
\hfill
\begin{adjustbox}{valign=t,minipage=0.60\linewidth}
\centering
\includegraphics[width=\linewidth]{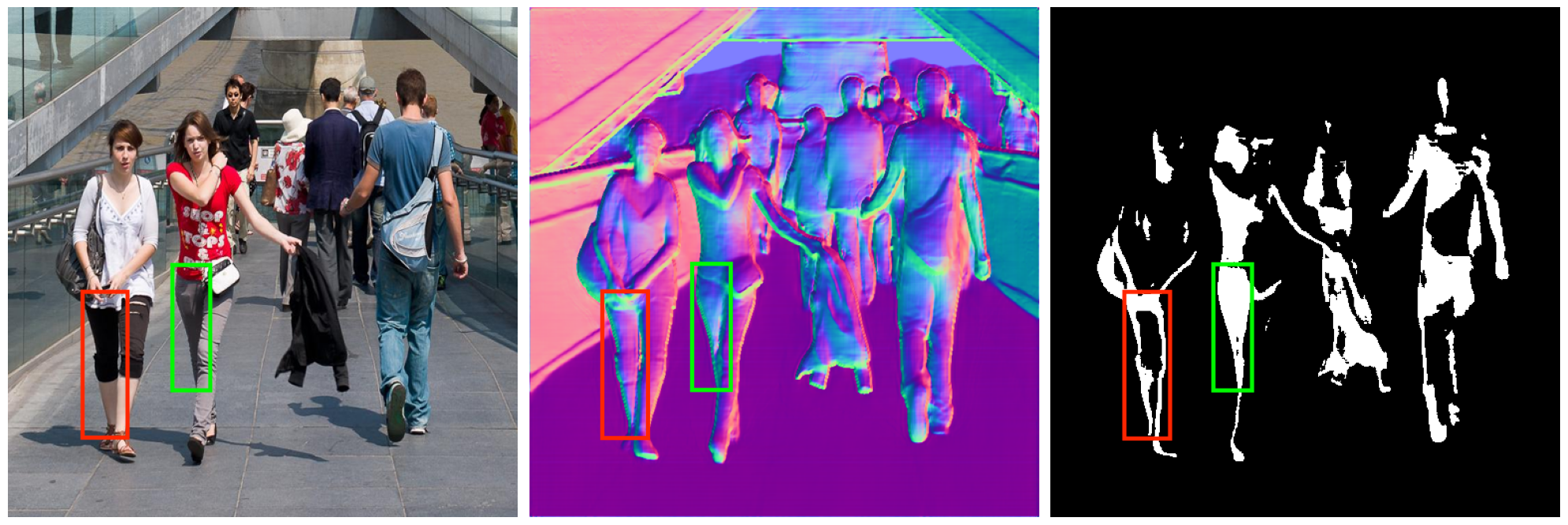}

{\small
  \makebox[0.33\linewidth]{Input Image}%
  \makebox[0.33\linewidth]{Normal Map}%
  \makebox[0.33\linewidth]{Ours Attached}%
}

\captionof{figure}{\textbf{Limitations.} Attached-shadow errors can arise from inaccurate surface normals, such as flipped normal directions that lead to incorrect attached-shadow predictions.}
\label{fig: limitation}

\end{adjustbox}

\end{figure}

Shadow masks are widely used in downstream tasks. To demonstrate the practical value of separating attached shadows, we study shadow removal as an example.
We guide an off-the-shelf shadow removal model \cite{guo2023shadowformer} using a \emph{joint} shadow mask, and then refine the result within the object region using the \emph{attached} shadow mask.
Because our data do not provide shadow-free ground truth for quantitative evaluation, we conduct a user study in which participants choose the more natural of the two outputs (see \cref{fig:user_study}).
As shown in \cref{tab: user-study}, participants prefer the result that uses the attached mask for refinement in 80.7\% of comparisons.
This outcome underscores the importance of separating cast and attached shadows: it enables targeted refinement on the object while removing cast shadows, leading to more realistic results.

\section{Limitations}
As an initial step toward detecting both cast and attached shadows, our framework has certain limitations. First, it relies on scene geometry derived from normal maps generated by an off-the-shelf model~\cite{yang2024depth}. However, this foundation model does not always yield accurate surface orientations, and such errors can propagate to our predictions, as illustrated in \cref{fig: limitation}. Improving the reliability of geometric cues through better normal estimation or geometry refinement is a promising direction for future work.
Second, our approach assumes a single directional light source, which may not generalize well to complex illumination settings, including multi-source indoor lighting, soft shadows from extended light sources, and low-light/night-time scenes with strong indirect illumination. Future work could extend the lighting formulation beyond a single directional source by modeling multiple or area lights, and broaden training data to cover more diverse illumination conditions.

\section{Conclusion}
We introduced a geometry- and light-aware framework for separately detecting cast and attached shadows, addressing the gap where most prior shadow detectors predict a unified mask or remain cast-focused.
Motivated by the relationship between illumination, surface geometry, and shadows, our method jointly learns shadow segmentation and light estimation: the estimated light direction and surface normals yield a geometry-derived partial attached shadow map, which is fed back to refine detection, while predicted shadows provide cues for updating the light direction.
This closed-loop, iterative formulation, together with explicit supervision for cast/attached separation, consistently improves attached-shadow detection while maintaining strong cast-shadow performance, as shown by comparisons to state-of-the-art baselines.
To support future research and model development, we introduce a new dataset with high-quality separate annotations for cast and attached shadows.

\section*{Acknowledgements}
This work was supported in part by the CCI startup fund at UNC Charlotte and NSF Grant IIS-2212046.

\clearpage

\title{Cast and Attached Shadow Detection via Iterative Light and Geometry Reasoning}
\subtitle{Supplementary Material}

\titlerunning{Attached Shadow Detection via Iterative Reasoning}
\author{
Shilin Hu\inst{1}\orcidlink{0009-0006-6002-1403}\and
Jingyi Xu\inst{1}\and
Sagnik Das\inst{1}\and
Dimitris Samaras\inst{1}\textsuperscript{$\dagger$}\orcidlink{0000-0002-1373-0294}\and
Hieu Le\inst{2}\textsuperscript{$\dagger$}\orcidlink{0000-0001-7855-2778}
}
\authorrunning{S.~Hu et al.}
\institute{
Stony Brook University, Stony Brook NY 11794, USA \and UNC Charlotte, Charlotte NC 28223, USA\\
\email{\{shilhu,jingyixu,sadas,samaras\}@cs.stonybrook.edu},
\email{hle40@charlotte.edu}
}

\maketitle
\begingroup
\renewcommand{\thefootnote}{}
\NoHyper
\footnotetext{\textsuperscript{$\dagger$} Equal advising.}
\endNoHyper
\endgroup

\setcounter{section}{0}
\setcounter{figure}{0}
\setcounter{table}{0}
\setcounter{equation}{0}

\noindent In this supplementary material, we provide the following:
\begin{enumerate}
    \item Additional visual results, including results for the \textbf{shadow removal application}, \textbf{cross-dataset} evaluation, and \textbf{qualitative} comparisons.
    \item Additional details on the framework design and evaluation, including the detailed \textbf{model architecture}, full \textbf{quantitative comparisons}, \textbf{computational overhead}, and \textbf{light estimation} performance.
    \item Details of \textbf{dataset acquisition and curation}.
    \item In addition to this PDF, we provide animated GIF examples illustrating cast and attached video shadow detection. Since these videos do not include attached shadow annotations, quantitative evaluation is not reported.
\end{enumerate}

\section{More Qualitative Results}
\begin{figure}[!b]
  \centering

  \begin{subfigure}[t]{0.495\linewidth}
    \centering
    \includegraphics[width=\linewidth]{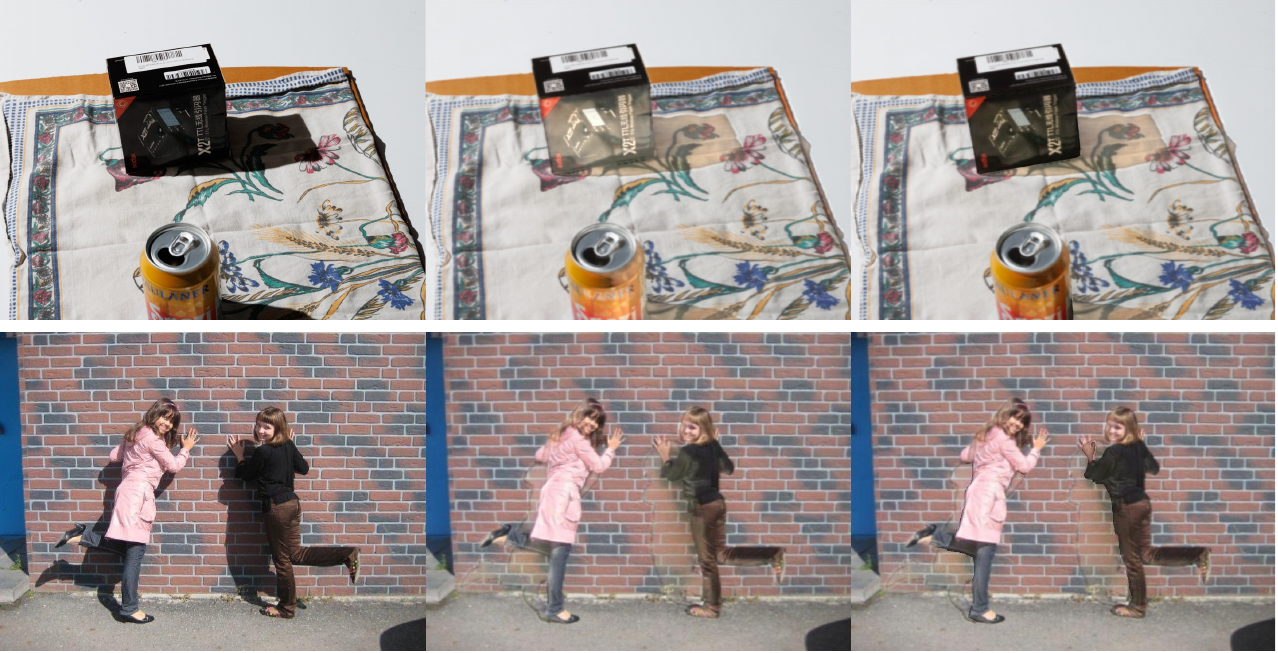}
    {\small
      \makebox[0.33\linewidth]{Input}%
      \makebox[0.33\linewidth]{Unified}%
      \makebox[0.33\linewidth]{Separate}%
    }
  \end{subfigure}\hfill
  \begin{subfigure}[t]{0.495\linewidth}
    \centering
    \includegraphics[width=\linewidth]{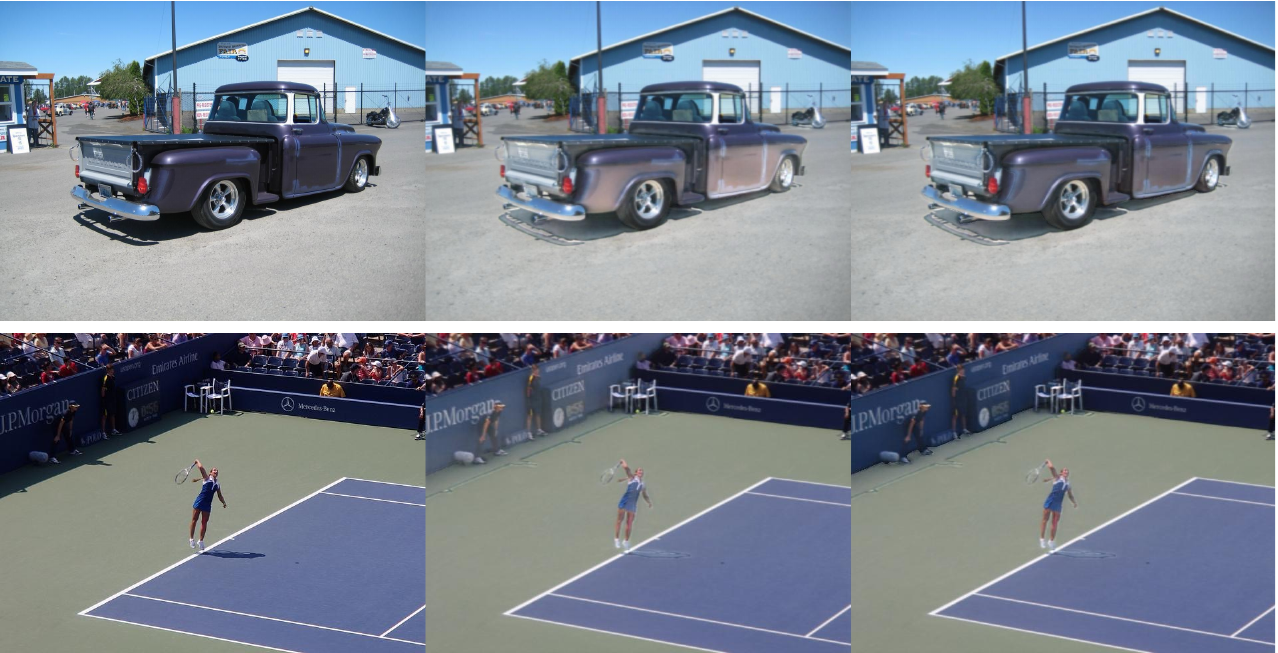}
    {\small
      \makebox[0.33\linewidth]{Input}%
      \makebox[0.33\linewidth]{Unified}%
      \makebox[0.33\linewidth]{Separate}%
    }
  \end{subfigure}

  \caption{Shadow removal application. Using a separate attached shadow mask for targeted refinement within the object region can produce more natural shadow removal.}
  \label{fig: removal}
\end{figure}

\paragraph{Shadow Removal.}
We conduct a user study to evaluate shadow removal performance. The study contains a pool of 40 questions. Each participant evaluates 20 randomly selected questions from this pool, and the presentation order of the unified and separate removal results is randomized. In total, we collected 259 votes, of which 209 favor the separate refinement over the unified refinement, corresponding to 80.7\%.

Additional results for shadow removal using ShadowFormer~\cite{guo2023shadowformer} with a unified shadow mask and separate cast and attached shadow masks are shown in \cref{fig: removal}. These examples demonstrate that using a separate attached shadow mask for refinement within the object region leads to more natural color recovery.

\begin{figure}[!t]
  \centering
  \begin{minipage}{0.8\linewidth}
  \includegraphics[width=\linewidth]{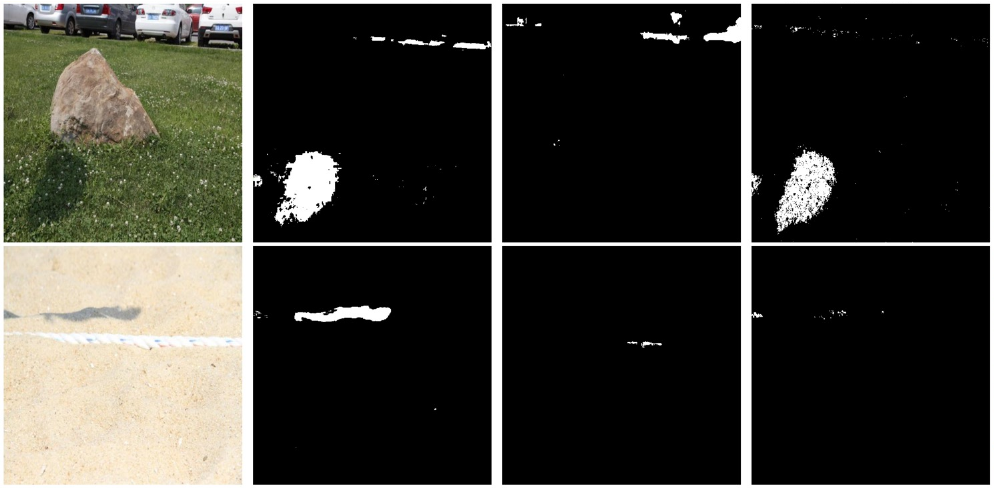}
  \noindent
  \makebox[0.25\linewidth][c]{\small{SRD}}%
  \makebox[0.25\linewidth][c]{\small{Pred Cast}}%
  \makebox[0.25\linewidth][c]{\small{Pred Att.}}%
  \makebox[0.25\linewidth][c]{\small{SRD GT}}%
  \end{minipage}
  \caption{Qualitative cross-dataset results on SRD \cite{Qu_2017_CVPR}, an image shadow detection dataset.}
  \label{fig: cross}
\end{figure}

\begin{figure}[!t]
  \centering
  \includegraphics[width=\linewidth]{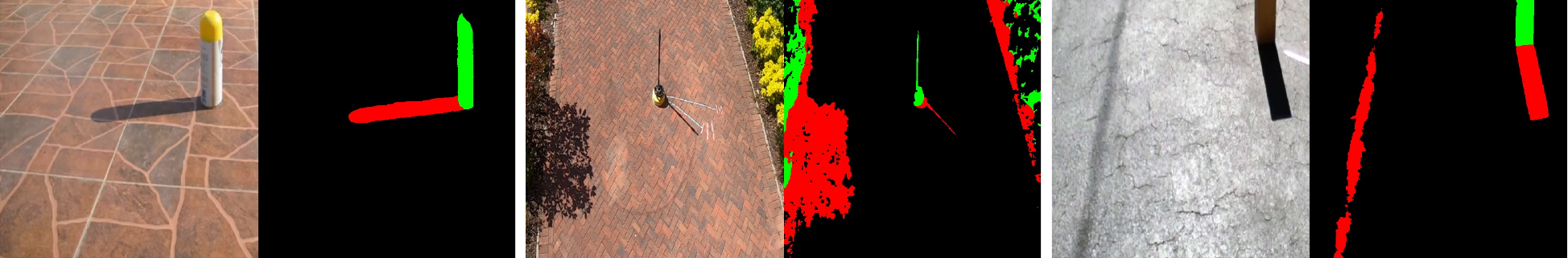}
  \caption{Qualitative cross-dataset results on SBU-TimeLapse \cite{le2021physics}, a video shadow detection dataset with static scenes.}
  \label{fig: timelapse}
\end{figure}

\begin{figure}[!t]
  \centering
  \includegraphics[width=\linewidth]{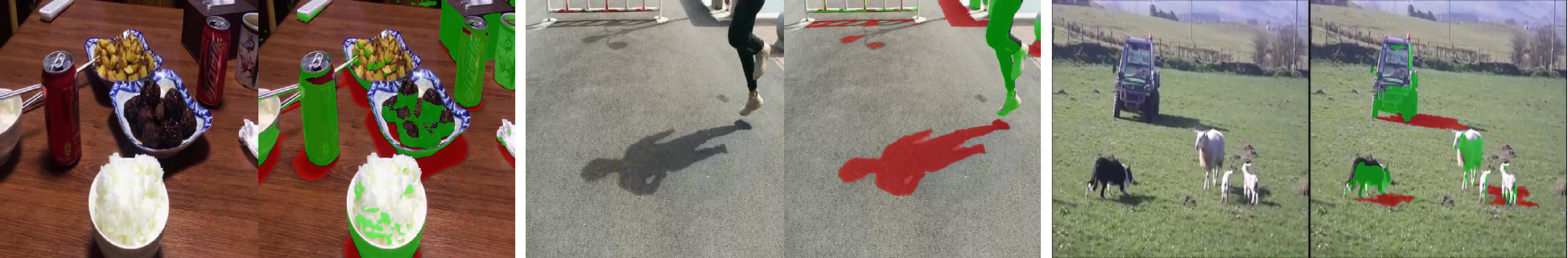}
  \caption{Qualitative cross-dataset results on ViSha \cite{Chen2021TriplecooperativeVS}, a video shadow detection dataset with dynamic scenes.}
  \label{fig: visha}
\end{figure}

\paragraph{Cross-dataset.}
Additional cross-dataset results are shown in \cref{fig: cross}, \cref{fig: timelapse}, and \cref{fig: visha}. We evaluate our method on the image shadow detection dataset SRD \cite{Qu_2017_CVPR}, as well as the video shadow detection datasets SBU-TimeLapse \cite{le2021physics} and ViSha \cite{Chen2021TriplecooperativeVS}. The results qualitatively show that our method generalizes across both image and video settings. 
Additional video results on SBU-TimeLapse \cite{le2021physics} and ViSha \cite{Chen2021TriplecooperativeVS} are provided in the \texttt{gif} folder for better visualization.

\begin{figure}[!t]
  \centering
  \includegraphics[width=\linewidth]{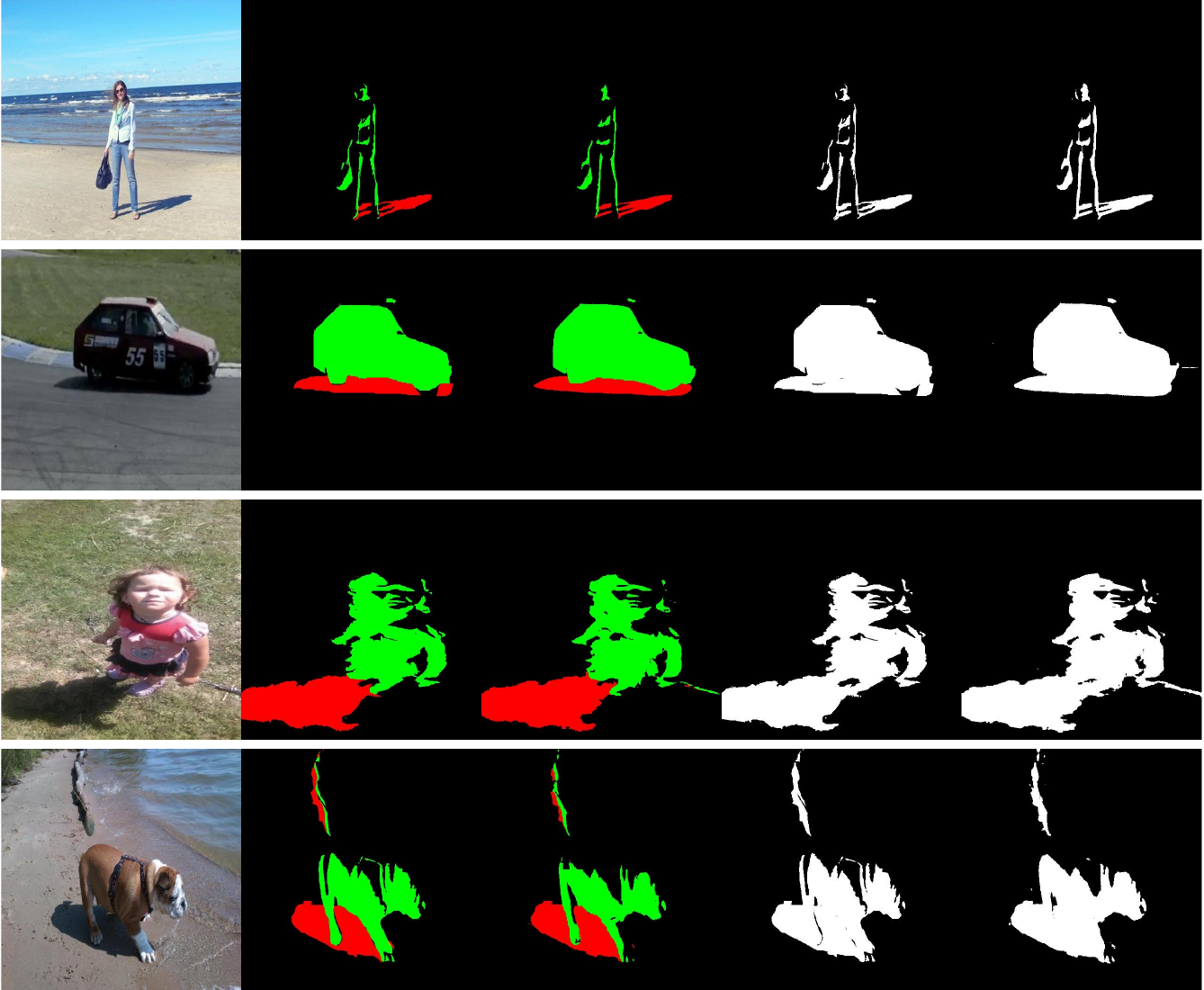}
  \makebox[0.19\linewidth]{\small{Image}}
  \makebox[0.19\linewidth]{\small{GT Cast \& Att.}}
  \makebox[0.19\linewidth]{\small{Pred Cast \& Att.}}
  \makebox[0.19\linewidth]{\small{GT Full}}
  \makebox[0.19\linewidth]{\small{Pred Full}}
  \caption{Additional qualitative examples of cast, attached, and full shadow predictions across diverse scenes in our dataset.}
  \label{fig: sup1}
\end{figure}

\begin{figure}[!t]
  \centering
  \includegraphics[width=\linewidth]{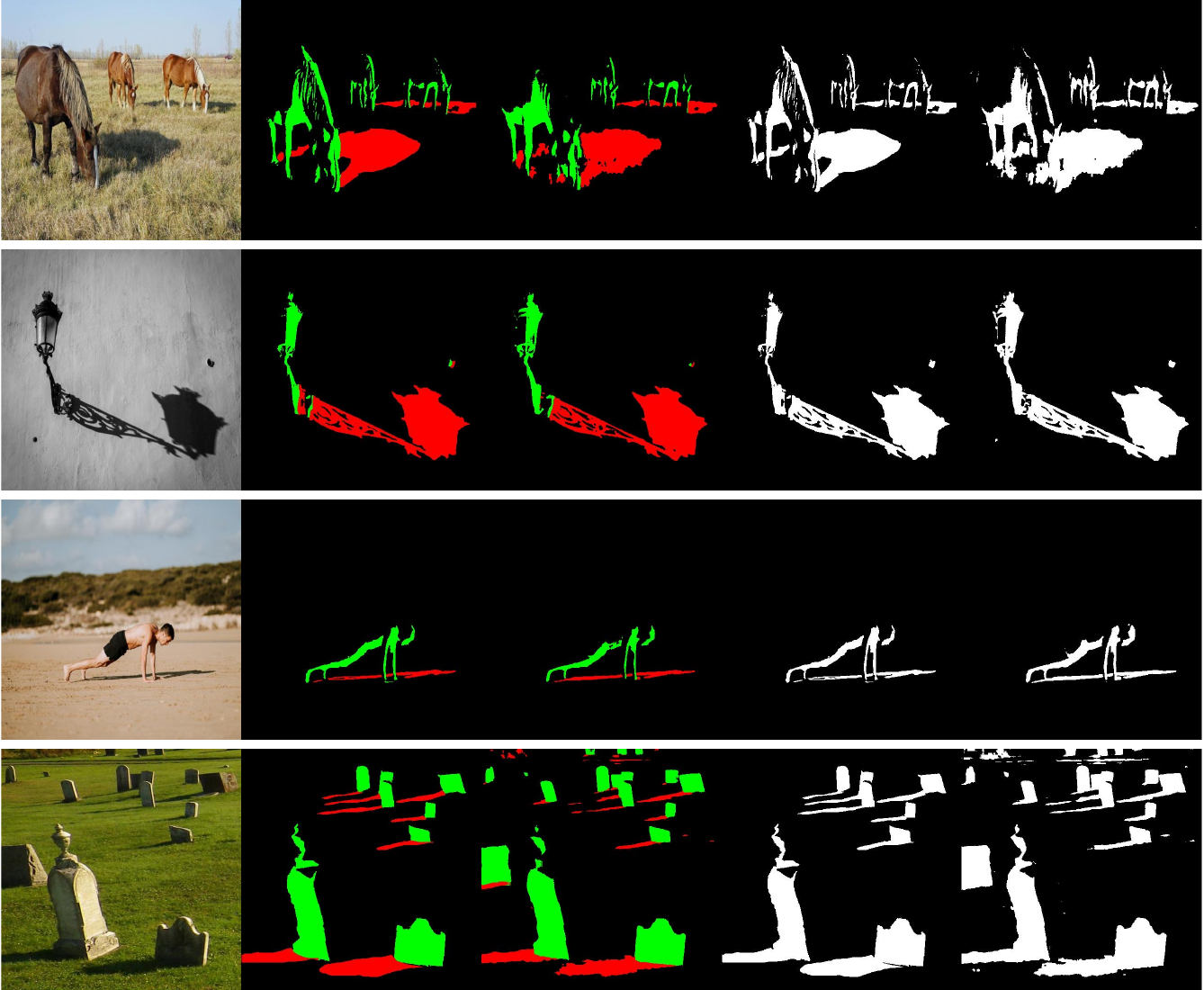}
  \makebox[0.19\linewidth]{\small{Image}}
  \makebox[0.19\linewidth]{\small{GT Cast \& Att.}}
  \makebox[0.19\linewidth]{\small{Pred Cast \& Att.}}
  \makebox[0.19\linewidth]{\small{GT Full}}
  \makebox[0.19\linewidth]{\small{Pred Full}}
  \caption{Additional qualitative examples of cast, attached, and full shadow predictions across diverse scenes in our dataset.}
  \label{fig: sup2}
\end{figure}

\begin{figure}[!t]
  \centering
  \includegraphics[width=\linewidth]{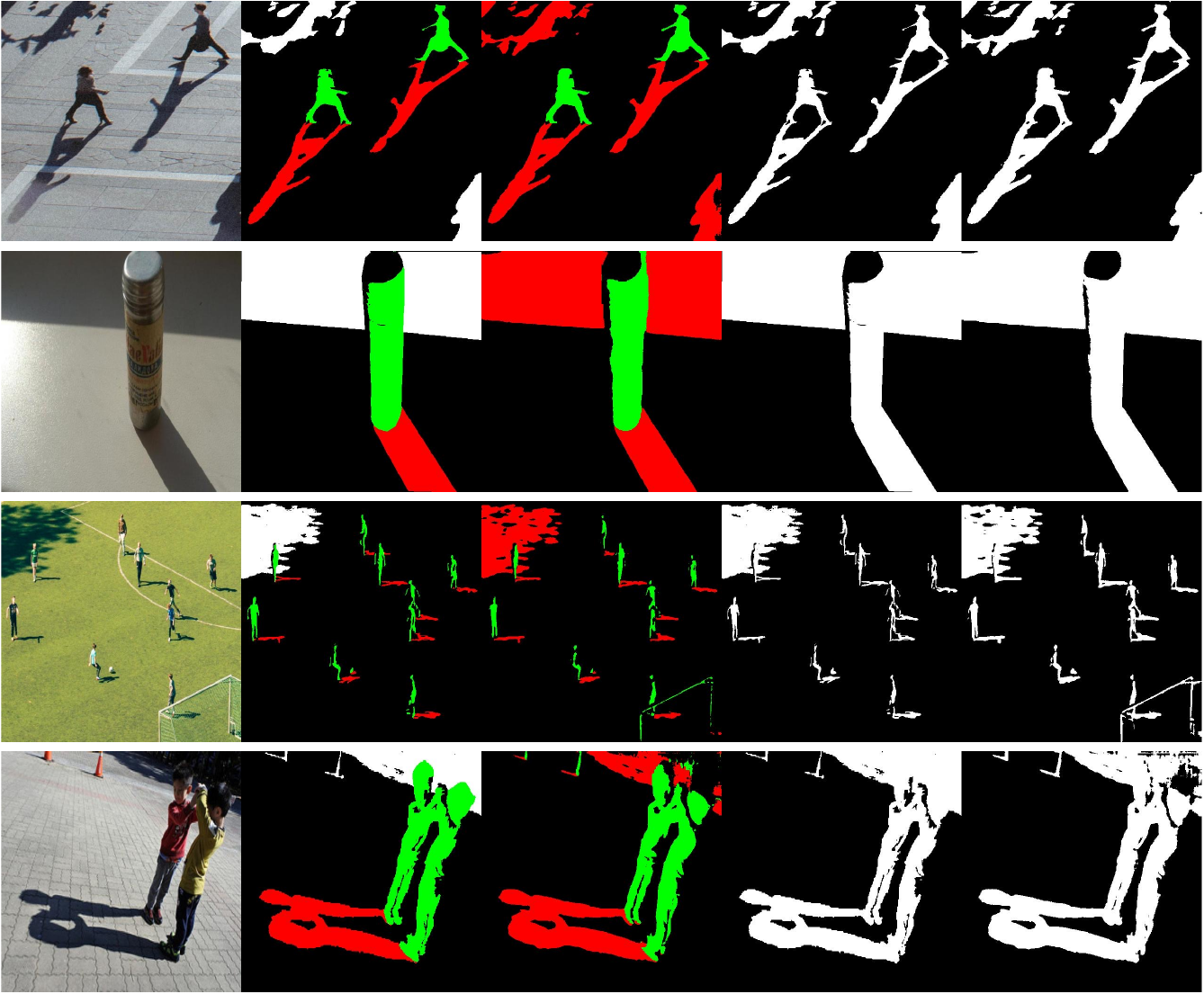}
  \makebox[0.19\linewidth]{\small{Image}}
  \makebox[0.19\linewidth]{\small{GT Anno.}}
  \makebox[0.19\linewidth]{\small{Pred Cast \& Att.}}
  \makebox[0.19\linewidth]{\small{GT Full}}
  \makebox[0.19\linewidth]{\small{Pred Full}}
  \caption{Additional qualitative examples of cast, attached, and full shadow predictions across diverse scenes in our dataset. Undefined shadows are marked as white in GT.}
  \label{fig: sup3}
\end{figure}

\paragraph{Qualitative.}
Additional qualitative results for cast, attached, and full shadow detection are shown in \cref{fig: sup1}, \cref{fig: sup2}, and \cref{fig: sup3}. Each figure displays the input image, ground-truth cast and attached shadow masks, our predicted cast and attached masks, and the corresponding ground-truth and predicted full shadow masks across diverse scenes.

\clearpage

\section{Framework Design and Evaluation}
We introduce a joint framework for shadow detection and light estimation, where the shadow detection module simultaneously predicts both cast and attached shadows.
All experiments are conducted on NVIDIA TITAN RTX GPUs. While baseline models run on a single GPU, our method requires two GPUs for training.
In this section, we detail the modifications to the network modules and provide additional evaluations on different shadow-detection configurations, light-estimation performance, and the computational overhead.

\subsection{Network Architecture}
The shadow detection module is adapted from the SILT model \cite{yang2023silt}, which features an encoder-decoder architecture with skip connections. The encoder is based on PVT-b5 \cite{wang2022pvt}, and we modify its first layer to incorporate additional input from the normal map and iterative feedback from the prior mask. The remaining encoder layers retain the original architecture, are initialized from SBU pre-trained weights, and are fine-tuned with a learning rate that is $0.05$ times that used for the rest of the framework. Only minimal changes are made to the decoder: we introduce a multi-channel segmentation head in the final layer with separate outputs for background, cast shadows, and attached shadows.

The light estimation module is adapted from ConvNeXt-S \cite{liu2022convnet}, with modifications to the input layer and the final output layer. The output of the light estimation module is a three-dimensional vector, where each dimension represents one component of the light direction.

For retraining prior methods \cite{zhu18b, Hu2019RevisitingSD, zhu2021mitigating, yang2023silt}, we use the full shadow mask (cast shadow + attached shadow + undefined shadow) as the supervision signal, as these methods formulate shadow detection as binary segmentation.

\begin{table}[!t]
\centering
\caption{Quantitative results for shadow detection methods. We report the balanced error rate (BER), Precision (P), Recall (R), and F-measure (F) for full, cast and attached shadow detection. $\dagger$ denotes retraining on our data using the full-shadow mask.}
\label{tab: allmetric}
\resizebox{\linewidth}{!}{%
\begin{tabular}{@{}lcccccccccccccccccc@{}}
\toprule
\multicolumn{1}{c}{\multirow{2}{*}{\textbf{Methods}}} & \multicolumn{6}{c}{\textbf{Full}} & \multicolumn{6}{c}{\textbf{Cast}} & \multicolumn{6}{c}{\textbf{Attached}} \\ \cmidrule(l){2-19} 
\multicolumn{1}{c}{} & \textbf{BER} & \textbf{S} & \textbf{NS} & \textbf{P} & \textbf{R} & \textbf{F1} & \textbf{BER} & \textbf{S} & \textbf{NS} & \textbf{P} & \textbf{R} & \textbf{F1} & \textbf{BER} & \textbf{S} & \textbf{NS} & \textbf{P} & \textbf{R} & \textbf{F1} \\ \midrule
BDRAR\cite{zhu18b} & 21.29 & 39.68 & 2.90 & 84.91 & 60.32 & 70.53 & 7.23 & 12.35 & 2.12 & 72.97 & 87.65 & 79.64 & 35.41 & 61.83 & 8.99 & 83.17 & 38.17 & 52.32 \\
FSDNet\cite{Hu2019RevisitingSD} & 24.46 & 46.83 & 2.09 & 87.30 & 53.17 & 66.09 & 10.51 & 19.53 & 1.49 & 77.89 & 80.47 & 79.16 & 37.53 & 68.14 & 6.92 & 84.27 & 31.86 & 46.23 \\
MTMT\cite{chen2020multi} & 33.10 & 64.15 & 2.05 & 82.59 & 35.85 & 50.00 & 15.16 & 29.07 & 1.25 & 78.81 & 70.93 & 74.66 & 42.02 & 74.55 & 9.50 & 75.71 & 25.45 & 38.10 \\
FDRNet\cite{zhu2021mitigating} & 23.69 & 42.77 & 4.60 & 77.10 & 57.23 & 65.69 & 8.95 & 14.41 & 3.49 & 61.61 & 85.59 & 71.65 & 34.82 & 58.18 & 11.45 & 80.95 & 41.82 & 55.15 \\
SDCM\cite{zhu2022single} & 24.27 & 46.58 & 1.95 & 88.10 & 53.43 & 66.51 & 7.88 & 14.48 & 1.28 & 81.42 & 85.52 & 83.42 & 36.83 & 65.16 & 8.50 & 82.66 & 34.84 & 49.02 \\
SILT\cite{yang2023silt} & 20.20 & 36.67 & 3.73 & 82.12 & 63.32 & 71.51 & 4.64 & 6.75 & 2.54 & 70.64 & 93.25 & 80.39 & 32.70 & 51.92 & 13.47 & 80.59 & 48.08 & 60.23 \\ \midrule
BDRAR$^\dagger$ & 8.15 & 9.51 & 6.79 & 78.30 & 90.49 & 83.95 & 5.18 & 6.20 & 4.16 & 59.57 & 93.80 & 72.86 & 20.75 & 11.10 & 30.40 & 77.28 & 88.91 & 82.69 \\
FSDNet$^\dagger$ & 10.17 & 17.11 & 3.22 & 87.43 & 82.89 & 85.10 & 5.82 & 9.66 & 1.97 & 74.96 & 90.34 & 81.94 & 19.49 & 23.64 & 15.34 & 85.27 & 76.36 & 80.57 \\
FDRNet$^\dagger$ & 8.62 & 9.96 & 7.28 & 76.98 & 90.04 & 83.00 & 4.65 & 5.10 & 4.20 & 59.67 & 94.90 & 73.27 & 23.11 & 9.90 & 36.32 & 74.27 & 90.10 & 81.42 \\
SILT$^\dagger$ & 6.51 & 3.11 & 9.91 & 72.56 & 96.89 & 82.98 & 4.52 & 3.39 & 5.64 & 52.84 & 96.61 & 68.31 & 26.57 & 3.36 & 49.78 & 69.31 & 96.64 & 80.72 \\ \midrule
BDRAR$^\dagger$+Normal & 7.44 & 7.89 & 6.99 & 78.09 & 92.11 & 84.52 & 4.63 & 4.96 & 4.29 & 59.14 & 95.04 & 72.91 & 19.70 & 8.19 & 31.21 & 77.38 & 91.81 & 83.98 \\
SILT$^\dagger$+Normal & 8.53 & 7.57 & 9.48 & 72.51 & 92.43 & 81.26 & 6.54 & 7.05 & 6.02 & 50.23 & 92.95 & 65.20 & 22.79 & 6.68 & 38.90 & 73.62 & 93.32 & 82.31 \\ \midrule
Ours No Normal & 7.60 & 10.14 & 5.07 & 81.46 & 89.86 & 85.46 & 7.82 & 13.74 & 1.90 & 73.44 & 86.26 & 79.33 & 20.87 & 29.25 & 12.49 & 84.62 & 70.75 & 77.07 \\
Ours Non-iterative & 7.57 & 10.52 & 4.61 & 82.78 & 89.48 & 85.99 & 6.14 & 11.47 & 0.81 & 86.98 & 88.53 & 87.75 & 15.10 & 20.50 & 9.70 & 88.84 & 79.50 & 83.91 \\
Ours No $\mathcal{L}_{\text{dist}}$ & 7.19 & 9.63 & 4.76 & 82.48 & 90.37 & 86.25 & 6.16 & 11.20 & 1.12 & 82.79 & 88.80 & 85.69 & 13.45 & 20.95 & 5.94 & 92.82 & 79.05 & 85.38 \\
Ours No $\mathcal{L}_{\text{att}}$ & 6.89 & 9.44 & 4.35 & 83.78 & 90.56 & 87.03 & 5.82 & 10.38 & 1.26 & 81.31 & 89.62 & 85.26 & 14.61 & 21.24 & 7.98 & 90.56 & 78.76 & 84.25 \\
Ours No $\mathcal{L}_{\text{dir}}$ & 7.50 & 10.69 & 4.31 & 83.70 & 89.31 & 86.42 & 4.97 & 8.49 & 1.45 & 79.41 & 91.51 & 85.03 & 15.26 & 21.62 & 8.90 & 89.54 & 78.38 & 83.59 \\
Ours 2 iterations & 6.63 & 9.30 & 3.96 & 85.02 & 90.70 & 87.77 & 5.41 & 9.73 & 1.09 & 83.50 & 90.27 & 86.75 & 13.22 & 20.68 & 5.77 & 93.03 & 79.32 & 85.63 \\
\rowcolor{lightgray}Ours 3 iterations & 6.50 & 7.92 & 5.09 & 81.75 & 92.08 & 86.61 & 5.04 & 8.72 & 1.36 & 80.34 & 91.28 & 85.46 & 12.92 & 16.37 & 9.48 & 89.55 & 83.63 & 86.49 \\
Ours 5 iterations & 6.11 & 7.23 & 4.99 & 82.15 & 92.77 & 87.14 & 5.83 & 10.46 & 1.20 & 81.93 & 89.54 & 85.57 & 13.51 & 17.97 & 9.05 & 89.80 & 82.03 & 85.74 \\ \bottomrule
\end{tabular}%
}
\end{table}

\subsection{More Shadow Detection Evaluation}
\cref{tab: allmetric} reports the full set of metrics for our comparisons with previous methods, retrained baselines, and different configurations of our proposed framework. We report the \textit{Precision}, \textit{Recall}, \textit{F1 score}, \textit{BER}, and error rate for shadow and non-shadow regions for full-shadow, cast-shadow, and attached-shadow detection.

\begin{figure}[!t]
  \centering
  \begin{minipage}[t]{0.56\linewidth}
    \vspace{-4pt}
    \centering
    \captionof{table}{Inference time for different configurations of our framework compared to the retrained SILT model \cite{yang2023silt}. Inference times are reported in seconds per image.}
    \begin{tabular}{@{}lc@{}}
    \toprule
    \multicolumn{1}{c}{\textbf{Method}} & \textbf{Inference (s)} \\ \midrule
    SILT$\dagger$            & 0.137 \\ \midrule
    Ours Non-Iterative       & 0.176 \\ \midrule
    Ours Iterative - 3 iters & 0.545 \\
    Ours Iterative - 2 iters & 0.347 \\
    Ours Iterative - 5 iters & 0.894 \\ \bottomrule
    \end{tabular}%
    \label{tab: overhead}
  \end{minipage}
  \hfill
  \begin{minipage}[t]{0.42\linewidth}
    \vspace{-4pt}
    \centering
    \captionof{table}{Quantitative results comparing the standalone light estimation module with the light estimation module integrated into our proposed framework.}
    \vspace{12pt}
    \begin{tabular}{lc}
    \hline
    \multicolumn{1}{c}{\textbf{Model}} & \textbf{Light MAE} \\ \hline
    Standalone                         & 0.034              \\ \hline
    Ours final 1st Iter.               & 0.061              \\
    Ours final 2nd Iter.               & 0.050              \\
    Ours final 3rd Iter.               & 0.049              \\ \hline
    \end{tabular}%
    \label{tab: light}
  \end{minipage}
\end{figure}

\subsection{Computational Overhead}
\cref{tab: overhead} summarizes the computational overhead across different configurations. In the non-iterative setting, our method introduces only a marginal increase of roughly $0.04$ seconds per image over the retrained SILT baseline, mainly because of the lightweight jointly learned light-estimation module. In the iterative setting, inference time scales predictably with the number of refinement steps. Our final configuration with three iterations achieves the best overall tradeoff between performance and efficiency, with an inference time of $0.545$ seconds per image.

\begin{figure}[!t]
  \centering
  \includegraphics[width=\linewidth]{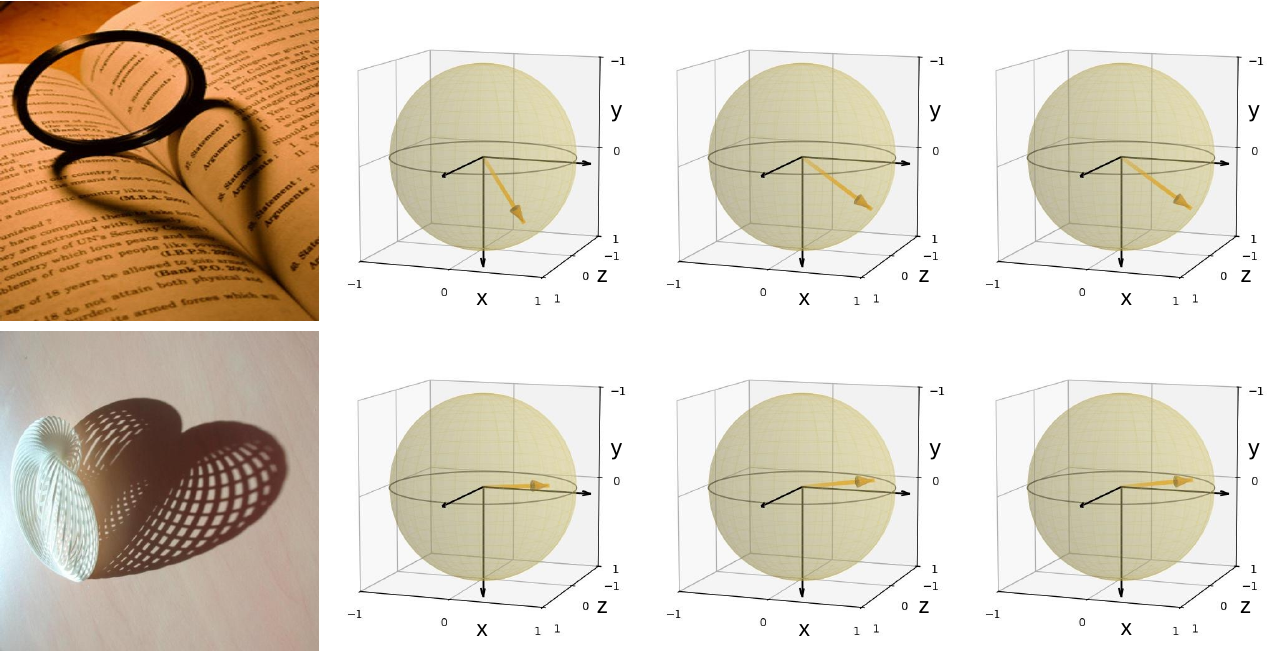}
  \makebox[0.24\linewidth]{\small{Image}}
  \makebox[0.24\linewidth]{\small{Iter 1}}
  \makebox[0.24\linewidth]{\small{Iter 2}}
  \makebox[0.24\linewidth]{\small{Iter 3}}
  \caption{Effectiveness of the iterative learning scheme. Our proposed feedback mechanism progressively refines the estimated light direction, with most of the improvement occurring by the second iteration and the results stabilizing at the third.}
  \label{fig: lightiter}
\end{figure}

\begin{figure}[!t]
  \centering
  \includegraphics[width=\linewidth]{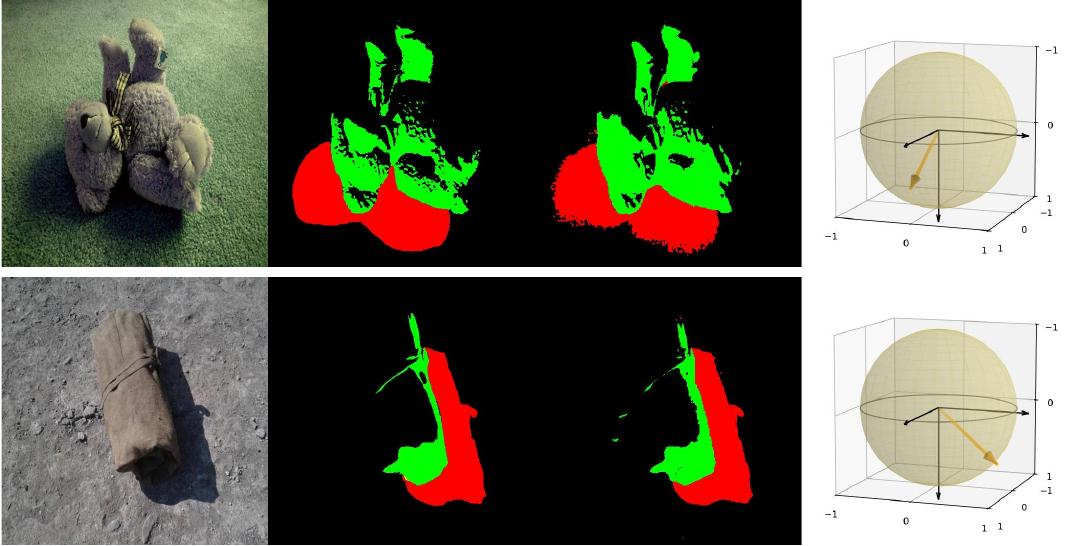}
  \makebox[0.24\linewidth]{\small{Image}}
  \makebox[0.24\linewidth]{\small{GT}}
  \makebox[0.24\linewidth]{\small{Prediction}}
  \makebox[0.24\linewidth]{\small{Light}}
  \caption{Qualitative examples of cast and attached shadow predictions obtained using our estimated light directions.}
  \label{fig: lightpred}
\end{figure}
\subsection{Standalone Light Estimation Performance}

To validate our light estimation strategy, we first evaluate a standalone light estimation module operating on ground-truth shadow masks. For SOBA and CUHK, heuristic light directions are used as pseudo ground truth, while WSRD~\cite{vasluianu2024ntire} provides physically calibrated lighting. As a result, light-direction accuracy is quantitatively evaluated only on the 42 WSRD test images. \cref{tab: light} compares this standalone estimator with the light predictor used in our joint framework.

With ground-truth shadow masks, the standalone module achieves a light-direction MAE of 0.034, serving as an approximate upper-bound reference under our evaluation setting. When the light predictor is integrated into the iterative framework and driven by predicted shadows, the MAE is 0.061 at the first iteration and improves to 0.050 and 0.049 at the second and third iterations, respectively. This progressive reduction indicates that the iterative scheme refines the light estimates and steadily approaches the upper-bound performance, as illustrated in \cref{fig: lightiter}. Qualitative examples in \cref{fig: lightpred} show that the predicted light directions are consistent with the observed cast and attached shadows.
\section{Dataset}
\label{sec: data}
We introduce the first dataset curated specifically for the joint detection of \emph{cast} and \emph{attached} shadows. The dataset contains 1,166 training images and 292 test images. Each image is annotated with a normal map, light direction, object mask, cast-shadow mask, attached-shadow mask, and undefined-shadow mask.

Our proposed dataset is constructed from three existing sources: WSRD~\cite{vasluianu2024ntire}, SOBA~\cite{wang2020instance}, and CUHK~\cite{Hu2019RevisitingSD}. These datasets were originally designed for different tasks: WSRD focuses on shadow removal with paired shadow/shadow-free images, SOBA provides instance-level cast-shadow annotations, and CUHK contains scene-level shadow masks. We unify and extend these annotations to support joint cast and attached shadow detection through the curation steps described below.

\begin{figure}[!b]
  \centering

  \begin{subfigure}[t]{0.495\linewidth}
    \centering
    \includegraphics[width=\linewidth]{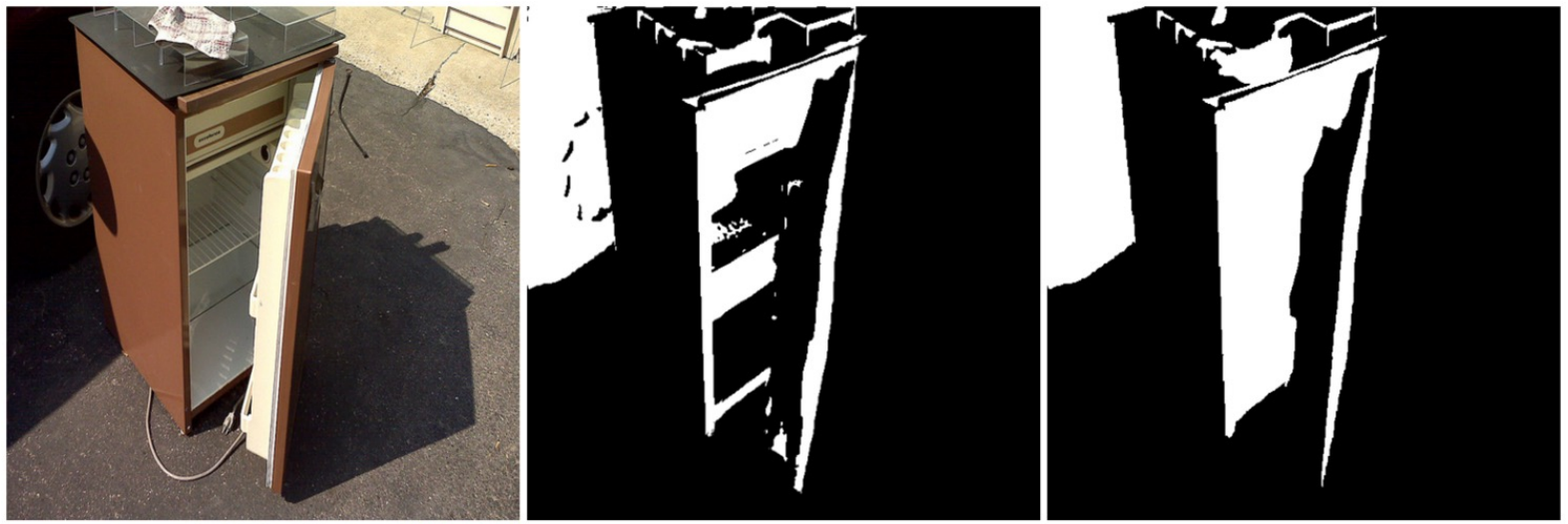}
    {\small
      \makebox[0.33\linewidth]{Input}%
      \makebox[0.33\linewidth]{Before}%
      \makebox[0.33\linewidth]{After}%
    }
  \end{subfigure}\hfill
  \begin{subfigure}[t]{0.495\linewidth}
    \centering
    \includegraphics[width=\linewidth]{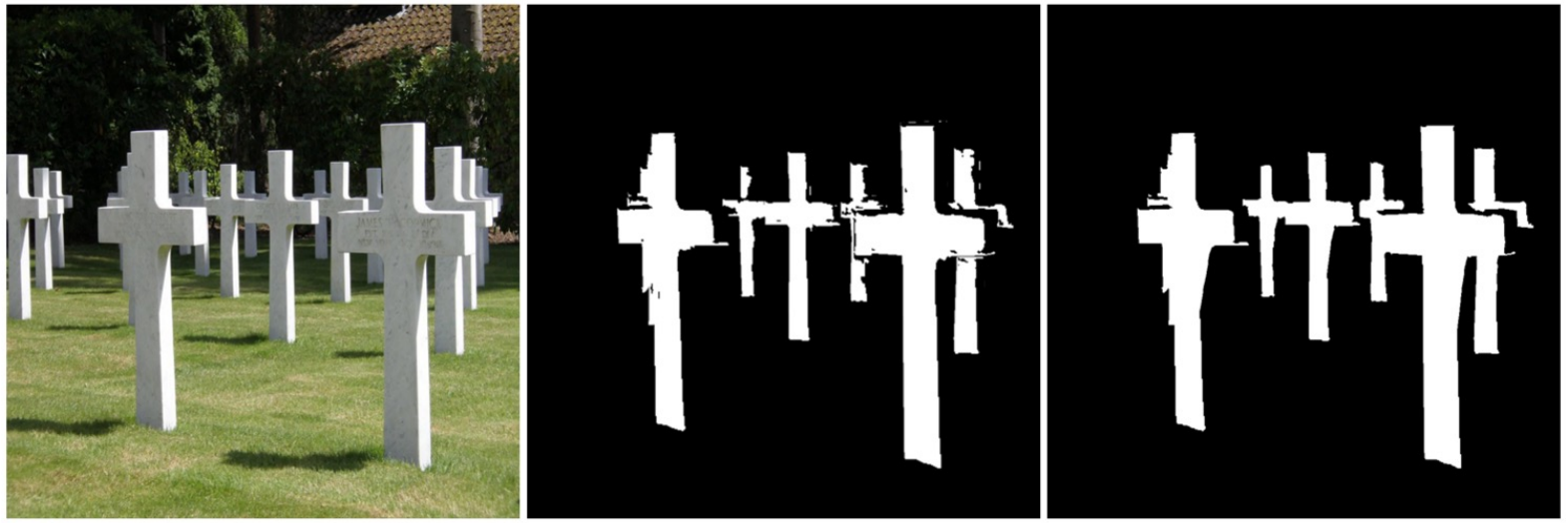}
    {\small
      \makebox[0.33\linewidth]{Input}%
      \makebox[0.33\linewidth]{Before}%
      \makebox[0.33\linewidth]{After}%
    }
  \end{subfigure}

  \caption{Examples of manual corrections applied to attached shadows in SOBA images.}
  \label{fig: associatesoba}
\end{figure}

\begin{figure}[!b]
  \centering

  \begin{subfigure}[t]{0.495\linewidth}
    \centering
    \includegraphics[width=\linewidth]{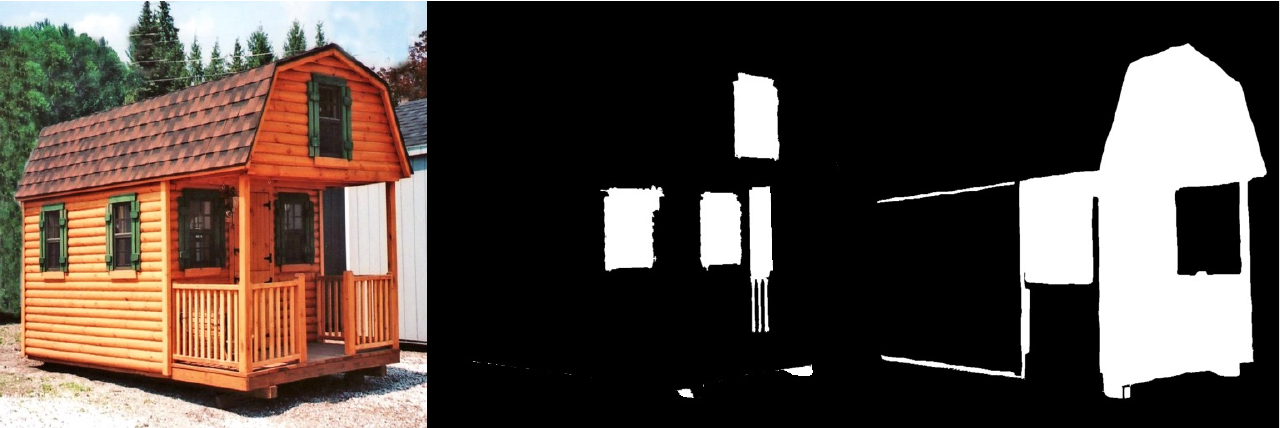}
    {\small
      \makebox[0.33\linewidth]{Input}%
      \makebox[0.33\linewidth]{Before}%
      \makebox[0.33\linewidth]{After}%
    }
  \end{subfigure}\hfill
  \begin{subfigure}[t]{0.495\linewidth}
    \centering
    \includegraphics[width=\linewidth]{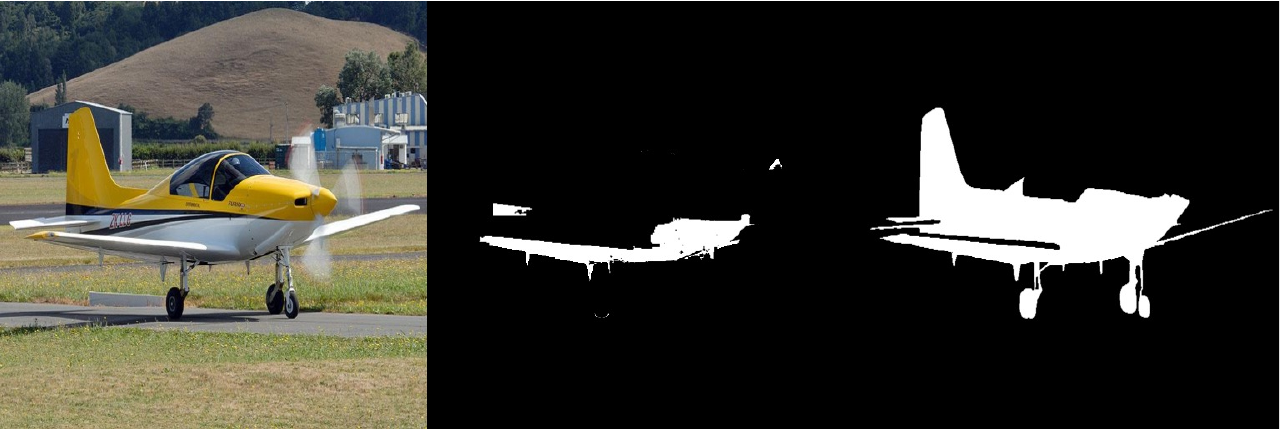}
    {\small
      \makebox[0.33\linewidth]{Input}%
      \makebox[0.33\linewidth]{Before}%
      \makebox[0.33\linewidth]{After}%
    }
  \end{subfigure}

  \caption{Examples of manual corrections applied to attached shadows in CUHK images.}
  \label{fig: associatecuhk}
\end{figure}

\begin{figure}[!t]
  \centering

  \begin{subfigure}[t]{0.495\linewidth}
    \centering
    \includegraphics[width=\linewidth]{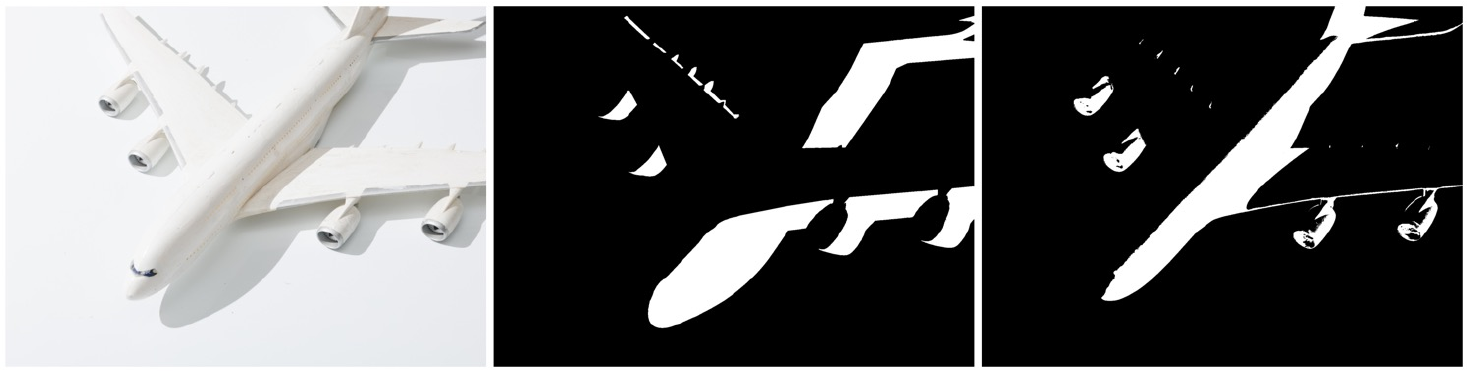}
    {\small
      \makebox[0.33\linewidth]{Input}%
      \makebox[0.33\linewidth]{Cast}%
      \makebox[0.33\linewidth]{Attached}%
    }
  \end{subfigure}\hfill
  \begin{subfigure}[t]{0.495\linewidth}
    \centering
    \includegraphics[width=\linewidth]{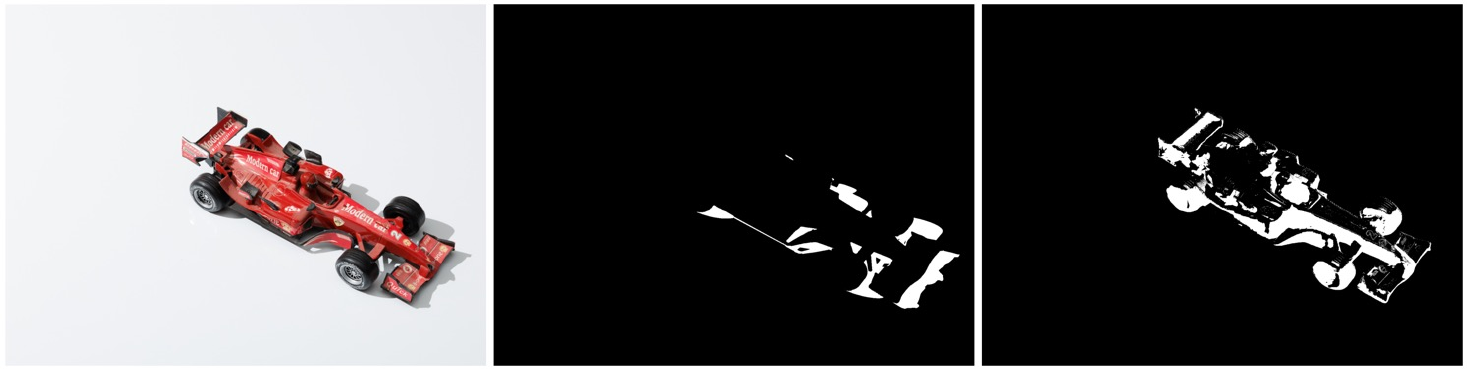}
    {\small
      \makebox[0.33\linewidth]{Input}%
      \makebox[0.33\linewidth]{Cast}%
      \makebox[0.33\linewidth]{Attached}%
    }
  \end{subfigure}

  \caption{Cast and attached shadow masks for WSRD images.}
  \label{fig: associatewsrd}
\end{figure}

\subsection{Cast and Attached Shadow Masks}
\paragraph{SOBA.}
Cast shadow masks are provided in the official SOBA annotations \cite{wang2020instance}. To generate attached shadow masks, we first use the derived light direction and precomputed normal map to create a prior attached-shadow mask. Errors in this prior mask are then manually corrected to obtain high-quality attached shadow annotations. Examples of these manual corrections are shown in \cref{fig: associatesoba}.
\paragraph{CUHK.}
CUHK provides full shadow masks for the scene; however, the quality of the attached shadow component in the provided annotations is not ideal, as attached shadows are not consistently labeled on object regions. Therefore, we manually correct the attached shadow masks for CUHK images. Examples of these corrections are shown in \cref{fig: associatecuhk}.
\paragraph{WSRD.}
For WSRD images, we obtain shadow masks by first performing color-based subtraction between the shadow and shadow-free images to identify regions with intensity changes. Next, we use the extracted object mask to differentiate between regions inside and outside the object, resulting in separate cast shadow masks and attached shadow masks, as illustrated in \cref{fig: associatewsrd}.

\subsection{Depth Map}
To enable reliable detection of attached shadows, accurate scene geometry is essential.
Depth estimation, a fundamental task in computer vision \cite{ranftl2020towards, ranftl2021vision, patni2024ecodepth, piccinelli2024unidepth, yang2024depthanything}, has recently demonstrated reliable performance on natural images.
We use the off-the-shelf Depth Anything V2~\cite{yang2024depth} to extract relative depth maps from our images.
The extracted depth maps are illustrated in \cref{fig: geo} (second column).

\subsection{Normal Map}
\begin{figure}[!t]
  \centering

  \begin{subfigure}[t]{0.495\linewidth}
    \centering
    \includegraphics[width=\linewidth]{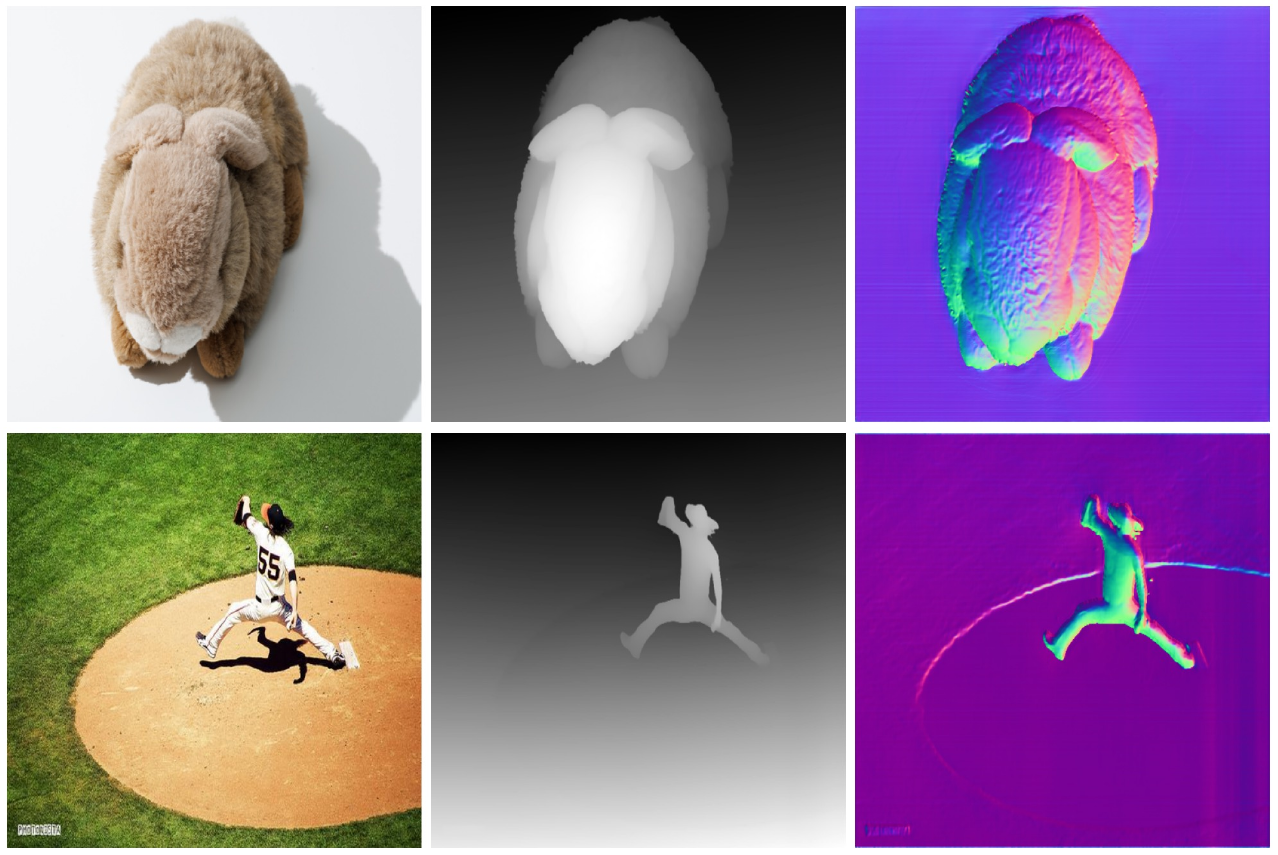}
    {\small
      \makebox[0.33\linewidth]{Input}%
      \makebox[0.33\linewidth]{Depth}%
      \makebox[0.33\linewidth]{Normal}%
    }
  \end{subfigure}\hfill
  \begin{subfigure}[t]{0.495\linewidth}
    \centering
    \includegraphics[width=\linewidth]{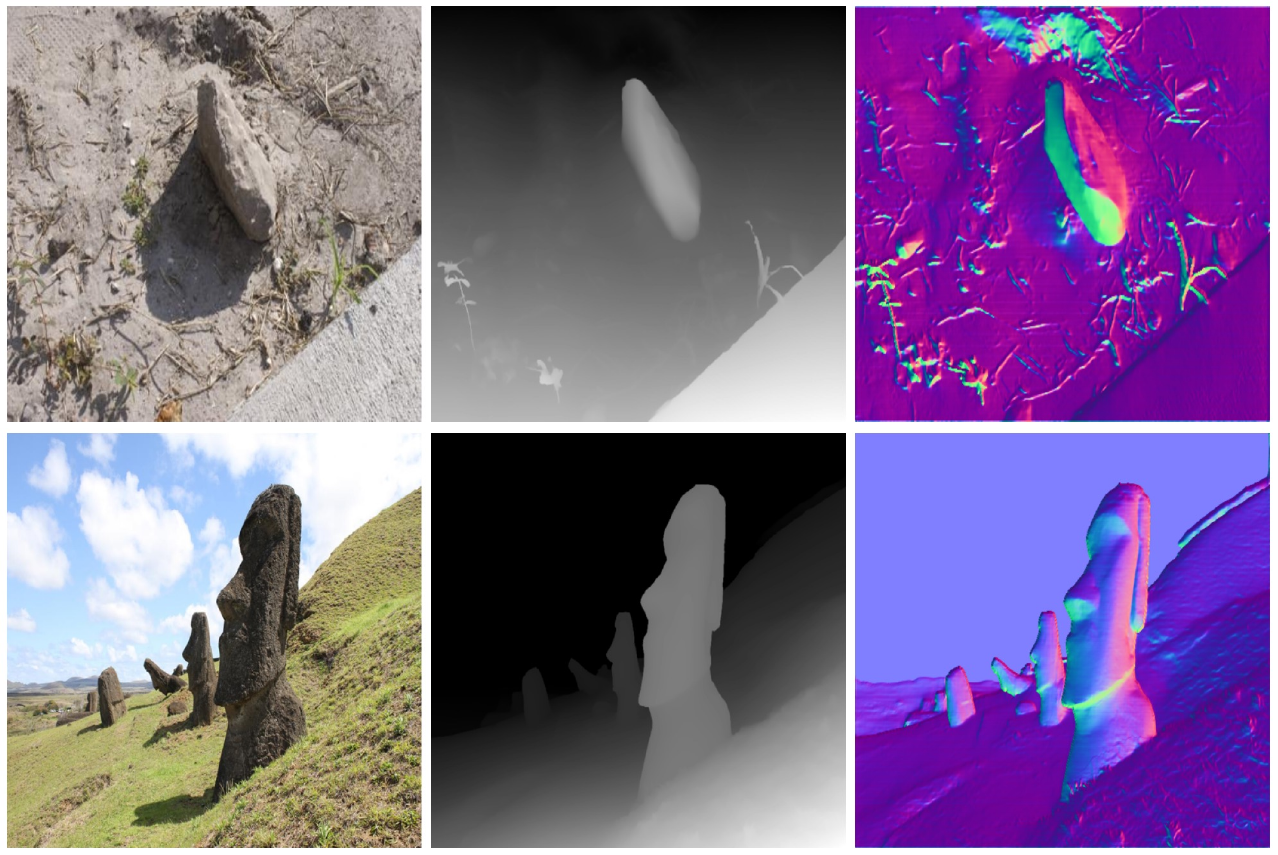}
    {\small
      \makebox[0.33\linewidth]{Input}%
      \makebox[0.33\linewidth]{Depth}%
      \makebox[0.33\linewidth]{Normal}%
    }
  \end{subfigure}

  \caption{Geometric information. We first extract depth maps using \cite{yang2024depth} and then derive the corresponding surface-normal maps relative to the camera viewpoint.}
  \label{fig: geo}
\end{figure}

Surface orientation plays a crucial role in determining attached shadows.
To capture this, we extract normal maps from the depth maps.
For each pixel, the third element (z-dimension) of the normal is set to 1, while the x and y components are given by the spatial gradients of the depth map.
Surface normals are computed relative to the camera viewpoint, with the positive z-dimension pointing inward.
The derived surface-normal maps are illustrated in \cref{fig: geo} (third column).

\begin{figure}[t]
  \centering

  \begin{minipage}[t]{0.40\linewidth}
    \vspace{0pt}
    \centering
    \includegraphics[height=5cm,width=\linewidth,keepaspectratio]{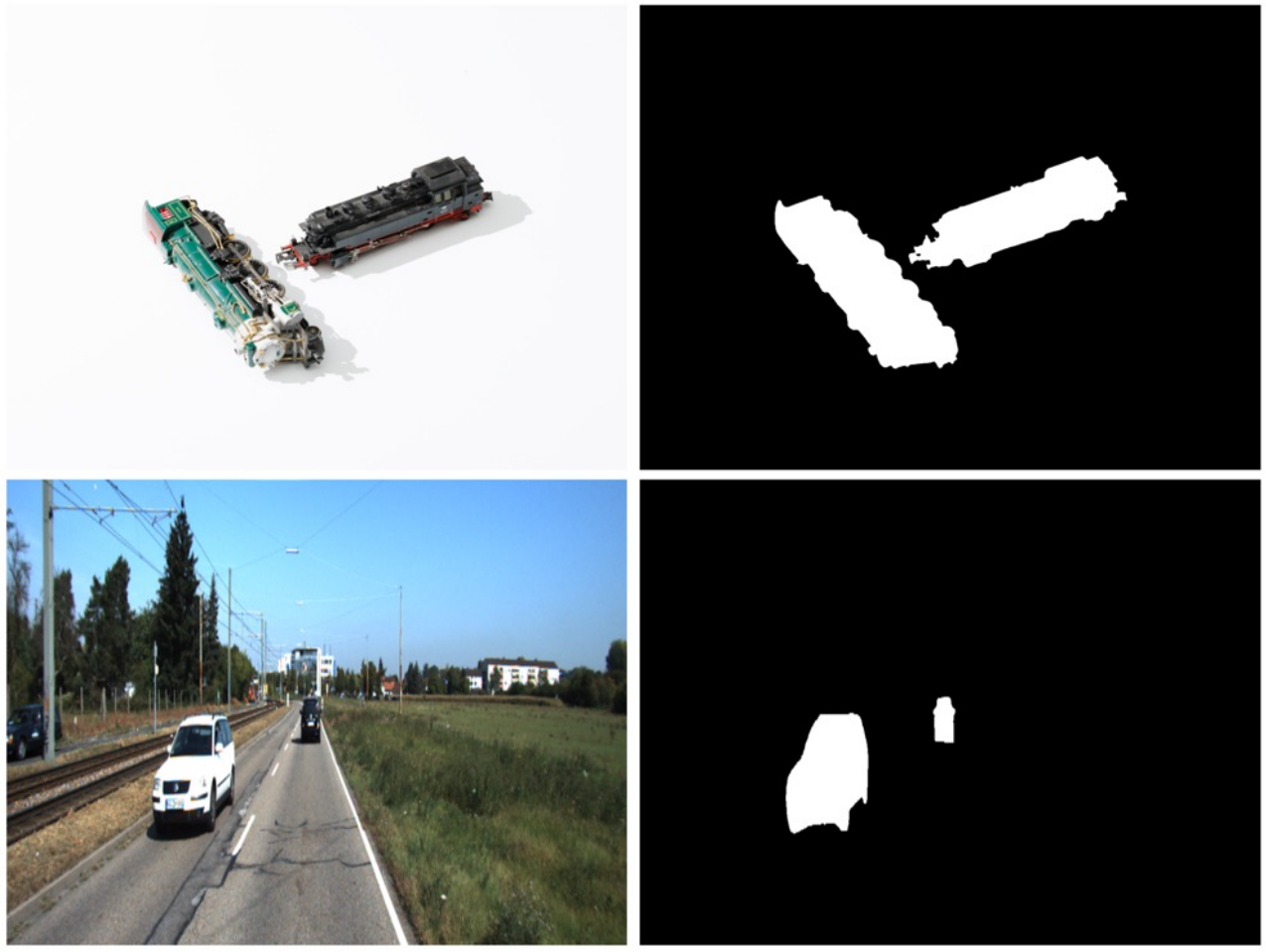}
    \vspace{-4pt}
    \caption{Extracted object masks for WSRD and CUHK images.}
    \label{fig: obj}
  \end{minipage}
  \hfill
  \begin{minipage}[t]{0.58\linewidth}
    \vspace{0pt}
    \centering
    \includegraphics[height=2.5cm,width=\linewidth,keepaspectratio]{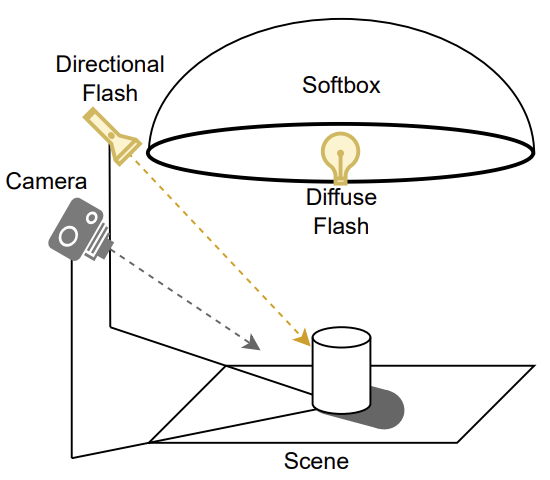}
    \vspace{-4pt}
    \caption{WSRD data-acquisition setup. The camera is positioned at a 45-degree vertical angle toward the scene. A directional light source is mounted at a 45-degree vertical angle toward the scene and at a 90-degree horizontal angle relative to the camera. Image adapted from~\cite{ntire2023wsrd}.}
    \label{fig: wsrd}
  \end{minipage}
\end{figure}

\subsection{Object Masks} 
For SOBA images, we directly use the object masks provided in the official annotations \cite{wang2020instance}. 
For WSRD and CUHK images, we use the state-of-the-art dichotomous image-segmentation model BiRefNet \cite{zheng2024bilateral} to automatically extract object masks from the input images.
These masks are used solely to define object-level annotations and for quantitative evaluation; they are not required by our method at test time.
\cref{fig: obj} shows representative examples of the extracted object masks for the WSRD and CUHK images.

\subsection{Light Direction}
Light direction is essential for identifying attached shadows.

For WSRD images, the calibrated acquisition setup (\cref{fig: wsrd}) provides known camera and light-source positions. Using the provided angles, we compute a 3D light-direction vector and express it in a camera-centric coordinate system ($+x$ right, $+y$ down, $+z$ inward).

\begin{figure}[!t]
  \centering
  \includegraphics[width=\linewidth]{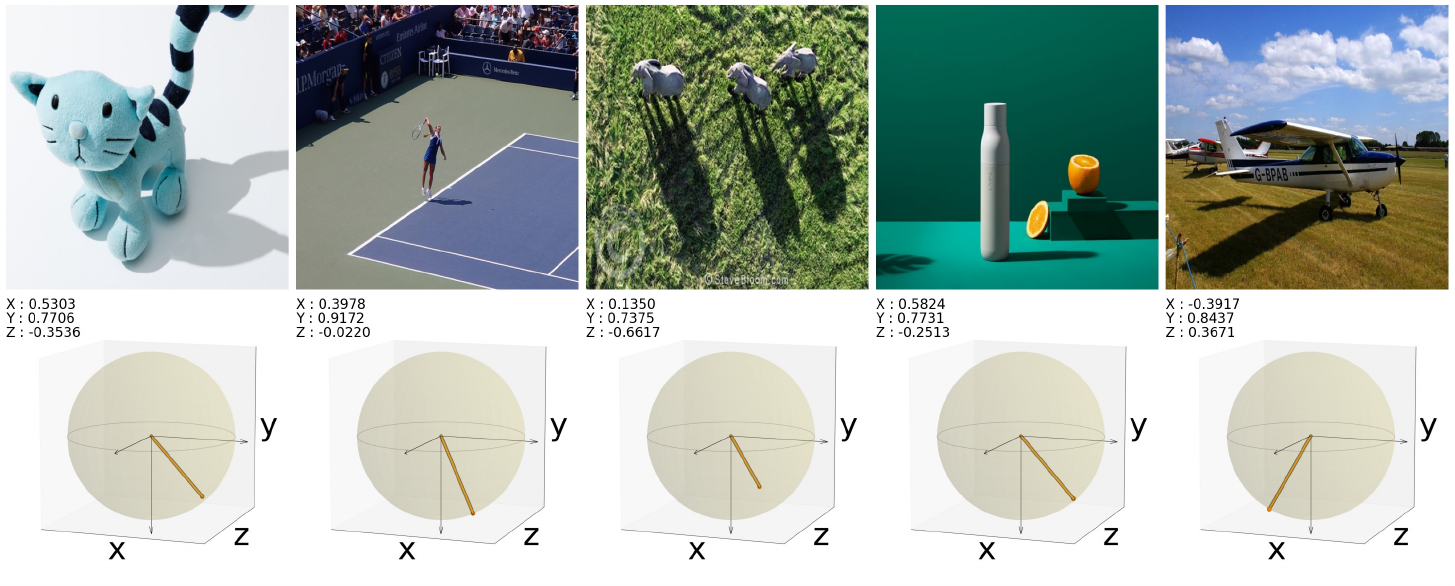}
  \caption{Derived light directions for WSRD, SOBA, and CUHK images, expressed in the camera-centric coordinate system. The $x$, $y$, and $z$ components of the unit light vectors are plotted.}
  \label{fig: lightdir}
\end{figure}

For SOBA and CUHK images, where calibration is unavailable, we estimate light direction from object--shadow pairs and depth. We first compute a 2D direction by connecting the centroid of an object to the centroid of its cast shadow, giving $(v_x, v_y)$. We then compute the average depth difference $\Delta d$ between object and shadow regions. The vector $(v_x, v_y, \Delta d)$ is normalized to obtain the light direction. When multiple object--shadow pairs exist, the resulting vectors are averaged to obtain a single direction for each image. The estimated light directions are illustrated in \cref{fig: lightdir}.

\begin{figure}[!t]
  \centering
  \includegraphics[width=0.95\linewidth]{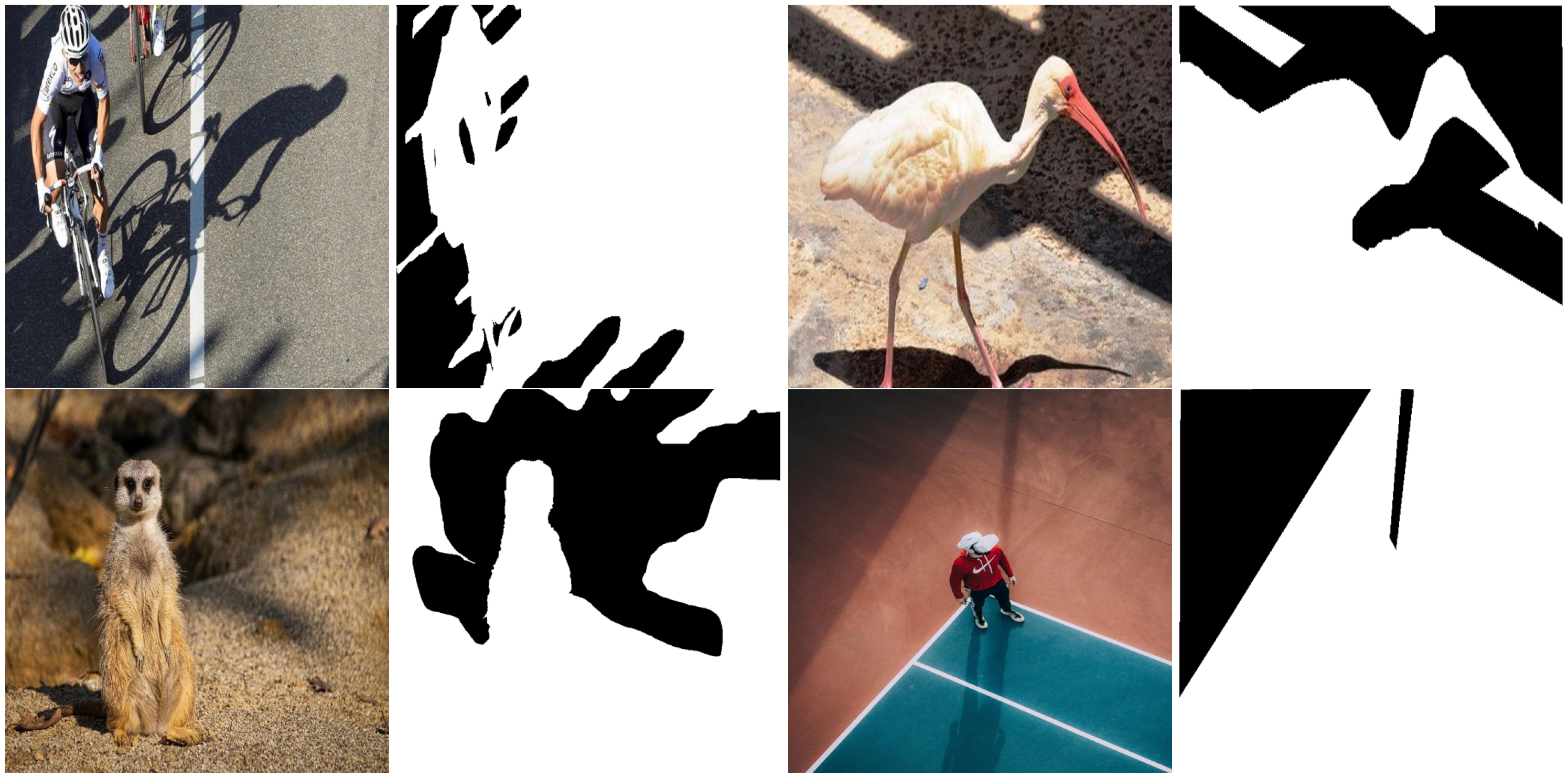}
  {\small
      \makebox[0.24\linewidth]{Input}%
      \makebox[0.24\linewidth]{Undefined}%
      \makebox[0.24\linewidth]{Input}%
      \makebox[0.24\linewidth]{Undefined}%
    }
  \caption{Illustration of excluding undefined shadow regions when training and evaluating cast- and attached-shadow predictions on our dataset. Undefined shadow masks are used only for training and evaluation of full-shadow predictions.}
  \label{fig: excluded}
\end{figure}

\subsection{Undefined Shadow Masks}
Our annotations focus on shadows associated with foreground objects. Shadow pixels that cannot be \textbf{reliably} attributed to these objects are labeled as \emph{undefined shadows}. 

For SOBA and CUHK images, we start from the provided scene-level shadow masks and mark regions not linked to annotated objects as undefined. These masks are used only for training and evaluation of full-shadow predictions. During cast/attached shadow training and evaluation, undefined regions are excluded (\cref{fig: excluded}).

\bibliographystyle{splncs04}
\bibliography{cites}

@misc{Adam,
  author        = {Diederik P. Kingma and Jimmy Ba},
  title         = {{Adam}: A Method for Stochastic Optimization},
  year          = {2017},
  doi           = {10.48550/arXiv.1412.6980},
  eprint        = {1412.6980},
  archivePrefix = {arXiv},
  primaryClass  = {cs.LG}
}

@article{bai2023local,
  author  = {Jiayang Bai and Zhen He and Shan Yang and Jie Guo and Zhenyu Chen and Yan Zhang and Yanwen Guo},
  title   = {Local-to-Global Panorama Inpainting for Locale-Aware Indoor Lighting Prediction},
  journal = {IEEE Trans. Vis. Comput. Graph.},
  volume  = {29},
  number  = {11},
  pages   = {4405--4416},
  year    = {2023},
  doi     = {10.1109/TVCG.2023.3320233}
}

@inproceedings{chen2020multi,
  author    = {Zhihao Chen and Lei Zhu and Liang Wan and Song Wang and Wei Feng and Pheng-Ann Heng},
  title     = {A Multi-Task Mean Teacher for Semi-Supervised Shadow Detection},
  booktitle = {Proc. IEEE/CVF Conf. Comput. Vis. Pattern Recog. (CVPR)},
  pages     = {5610--5619},
  year      = {2020},
  doi       = {10.1109/CVPR42600.2020.00565}
}

@inproceedings{Chen2021TriplecooperativeVS,
  author    = {Zhihao Chen and Liang Wan and Lei Zhu and Jia Shen and Huazhu Fu and Wennan Liu and Jing Qin},
  title     = {Triple-Cooperative Video Shadow Detection},
  booktitle = {Proc. IEEE/CVF Conf. Comput. Vis. Pattern Recog. (CVPR)},
  pages     = {2714--2723},
  year      = {2021},
  doi       = {10.1109/CVPR46437.2021.00274}
}

@inproceedings{chen2023sam,
  author    = {Tianrun Chen and Lanyun Zhu and Chaotao Ding and Runlong Cao and Yan Wang and Shangzhan Zhang and Zejian Li and Lingyun Sun and Ying Zang and Papa Mao},
  title     = {{SAM}-Adapter: Adapting Segment Anything in Underperformed Scenes},
  booktitle = {Proc. IEEE/CVF Int. Conf. Comput. Vis. Workshops (ICCVW)},
  pages     = {3359--3367},
  year      = {2023},
  doi       = {10.1109/ICCVW60793.2023.00361}
}

@article{Chuang2003,
  author  = {Yung-Yu Chuang and Dan B. Goldman and Brian Curless and David H. Salesin and Richard Szeliski},
  title   = {Shadow Matting and Compositing},
  journal = {ACM Trans. Graph.},
  volume  = {22},
  number  = {3},
  pages   = {494--500},
  year    = {2003},
  doi     = {10.1145/1201775.882298}
}

@inproceedings{Cun2020TowardsGS,
  author    = {Xiaodong Cun and Chi-Man Pun and Cheng Shi},
  title     = {Towards Ghost-Free Shadow Removal via Dual Hierarchical Aggregation Network and Shadow Matting {GAN}},
  booktitle = {Proc. AAAI Conf. Artif. Intell.},
  volume    = {34},
  pages     = {10680--10687},
  year      = {2020},
  doi       = {10.1609/aaai.v34i07.6695}
}

@inproceedings{dastjerdi2023everlight,
  author    = {Mohammad Reza Karimi Dastjerdi and Jonathan Eisenmann and Yannick Hold-Geoffroy and Jean-Fran{\c{c}}ois Lalonde},
  title     = {{EverLight}: Indoor-Outdoor Editable {HDR} Lighting Estimation},
  booktitle = {Proc. IEEE/CVF Int. Conf. Comput. Vis. (ICCV)},
  pages     = {7386--7395},
  year      = {2023},
  doi       = {10.1109/ICCV51070.2023.00682}
}

@inproceedings{Ding2019ARGANAR,
  author    = {Bin Ding and Chengjiang Long and Ling Zhang and Chunxia Xiao},
  title     = {{ARGAN}: Attentive Recurrent Generative Adversarial Network for Shadow Detection and Removal},
  booktitle = {Proc. IEEE/CVF Int. Conf. Comput. Vis. (ICCV)},
  pages     = {10212--10221},
  year      = {2019},
  doi       = {10.1109/ICCV.2019.01031}
}

@inproceedings{gardner2019deep,
  author    = {Marc-Andr{\'e} Gardner and Yannick Hold-Geoffroy and Kalyan Sunkavalli and Christian Gagn{\'e} and Jean-Fran{\c{c}}ois Lalonde},
  title     = {Deep Parametric Indoor Lighting Estimation},
  booktitle = {Proc. IEEE/CVF Int. Conf. Comput. Vis. (ICCV)},
  pages     = {7174--7182},
  year      = {2019},
  doi       = {10.1109/ICCV.2019.00727}
}

@inproceedings{guo2023shadowdiffusion,
  author    = {Lanqing Guo and Chong Wang and Wenhan Yang and Siyu Huang and Yufei Wang and Hanspeter Pfister and Bihan Wen},
  title     = {{ShadowDiffusion}: When Degradation Prior Meets Diffusion Model for Shadow Removal},
  booktitle = {Proc. IEEE/CVF Conf. Comput. Vis. Pattern Recog. (CVPR)},
  pages     = {14049--14058},
  year      = {2023},
  doi       = {10.1109/CVPR52729.2023.01350}
}

@inproceedings{guo2023shadowformer,
  author    = {Lanqing Guo and Siyu Huang and Ding Liu and Hao Cheng and Bihan Wen},
  title     = {{ShadowFormer}: Global Context Helps Shadow Removal},
  booktitle = {Proc. AAAI Conf. Artif. Intell.},
  volume    = {37},
  pages     = {710--718},
  year      = {2023},
  doi       = {10.1609/aaai.v37i1.25148}
}

@inproceedings{Hu_2018_CVPR,
  author    = {Xiaowei Hu and Lei Zhu and Chi-Wing Fu and Jing Qin and Pheng-Ann Heng},
  title     = {Direction-Aware Spatial Context Features for Shadow Detection},
  booktitle = {Proc. IEEE/CVF Conf. Comput. Vis. Pattern Recog. (CVPR)},
  pages     = {7454--7462},
  year      = {2018},
  doi       = {10.1109/CVPR.2018.00778}
}

@inproceedings{hu_iccv2019mask,
  author    = {Xiaowei Hu and Yitong Jiang and Chi-Wing Fu and Pheng-Ann Heng},
  title     = {{Mask-ShadowGAN}: Learning to Remove Shadows from Unpaired Data},
  booktitle = {Proc. IEEE/CVF Int. Conf. Comput. Vis. (ICCV)},
  pages     = {2472--2481},
  year      = {2019},
  doi       = {10.1109/ICCV.2019.00256}
}

@article{Hu2019RevisitingSD,
  author  = {Xiaowei Hu and Tianyu Wang and Chi-Wing Fu and Yitong Jiang and Qiong Wang and Pheng-Ann Heng},
  title   = {Revisiting Shadow Detection: A New Benchmark Dataset for Complex World},
  journal = {IEEE Trans. Image Process.},
  volume  = {30},
  pages   = {1925--1934},
  year    = {2021},
  doi     = {10.1109/TIP.2021.3049331}
}

@inproceedings{Hu2024ShadowRR,
  author    = {Shilin Hu and Hieu Le and ShahRukh Athar and Sagnik Das and Dimitris Samaras},
  title     = {Shadow Removal Refinement via Material-Consistent Shadow Edges},
  booktitle = {Proc. IEEE/CVF Winter Conf. Appl. Comput. Vis. (WACV)},
  pages     = {2631--2641},
  year      = {2025},
  doi       = {10.1109/WACV61041.2025.00261}
}

@inproceedings{jin2024des3,
  author    = {Yeying Jin and Wei Ye and Wenhan Yang and Yuan Yuan and Robby T. Tan},
  title     = {{DeS3}: Adaptive Attention-Driven Self and Soft Shadow Removal Using {ViT} Similarity},
  booktitle = {Proc. AAAI Conf. Artif. Intell.},
  volume    = {38},
  pages     = {2634--2642},
  year      = {2024},
  doi       = {10.1609/aaai.v38i3.28041}
}

@inproceedings{JingyiICCV21,
  author    = {Jingyi Xu and Hieu Le and Mingzhen Huang and ShahRukh Athar and Dimitris Samaras},
  title     = {Variational Feature Disentangling for Fine-Grained Few-Shot Classification},
  booktitle = {Proc. IEEE/CVF Int. Conf. Comput. Vis. (ICCV)},
  pages     = {8792--8801},
  year      = {2021},
  doi       = {10.1109/ICCV48922.2021.00869}
}

@inproceedings{Lalonde10,
  author    = {Jean-Fran{\c{c}}ois Lalonde and Alexei A. Efros and Srinivasa G. Narasimhan},
  title     = {Detecting Ground Shadows in Outdoor Consumer Photographs},
  booktitle = {Proc. Eur. Conf. Comput. Vis. (ECCV)},
  pages     = {322--335},
  year      = {2010},
  doi       = {10.1007/978-3-642-15552-9_24}
}

@inproceedings{Le_2019_CVPR_Workshops,
  author    = {Hieu M. Le and Bento Goncalves and Dimitris Samaras and Heather Lynch},
  title     = {Weakly Labeling the Antarctic: The Penguin Colony Case},
  booktitle = {Proc. IEEE/CVF Conf. Comput. Vis. Pattern Recog. Workshops (CVPRW)},
  pages     = {18--25},
  year      = {2019},
  doi       = {10.48550/arXiv.1905.03313}
}

@inproceedings{Le_2020_ECCV,
  author    = {Hieu Le and Dimitris Samaras},
  title     = {From Shadow Segmentation to Shadow Removal},
  booktitle = {Proc. Eur. Conf. Comput. Vis. (ECCV)},
  pages     = {264--281},
  year      = {2020},
  doi       = {10.1007/978-3-030-58621-8_16}
}

@article{Le_RS22,
  author  = {Hieu Le and Dimitris Samaras and Heather J. Lynch},
  title   = {A Convolutional Neural Network Architecture Designed for the Automated Survey of Seabird Colonies},
  journal = {Remote Sens. Ecol. Conserv.},
  volume  = {8},
  number  = {2},
  pages   = {251--262},
  year    = {2022},
  doi     = {10.1002/rse2.240}
}

@inproceedings{Le2016GeodesicDH,
  author    = {Hieu Le and Vu Nguyen and Chen-Ping Yu and Dimitris Samaras},
  title     = {Geodesic Distance Histogram Feature for Video Segmentation},
  booktitle = {Proc. Asian Conf. Comput. Vis. (ACCV)},
  pages     = {275--290},
  year      = {2016},
  doi       = {10.1007/978-3-319-54181-5_18}
}

@article{le2021physics,
  author  = {Hieu Le and Dimitris Samaras},
  title   = {Physics-Based Shadow Image Decomposition for Shadow Removal},
  journal = {IEEE Trans. Pattern Anal. Mach. Intell.},
  volume  = {44},
  number  = {12},
  pages   = {9088--9101},
  year    = {2022},
  doi     = {10.1109/TPAMI.2021.3124934}
}

@inproceedings{Le-etal-ICCV19,
  author    = {Hieu Le and Dimitris Samaras},
  title     = {Shadow Removal via Shadow Image Decomposition},
  booktitle = {Proc. IEEE/CVF Int. Conf. Comput. Vis. (ICCV)},
  pages     = {8577--8586},
  year      = {2019},
  doi       = {10.1109/ICCV.2019.00867}
}

@inproceedings{LeICCV2017,
  author    = {Hieu Le and Chen-Ping Yu and Gregory Zelinsky and Dimitris Samaras},
  title     = {Co-Localization with Category-Consistent Features and Geodesic Distance Propagation},
  booktitle = {Proc. IEEE Int. Conf. Comput. Vis. Workshops (ICCVW)},
  pages     = {1103--1112},
  year      = {2017},
  doi       = {10.1109/ICCVW.2017.134}
}

@inproceedings{li2023leveraging,
  author    = {Xiaoguang Li and Qing Guo and Rabab Abdelfattah and Di Lin and Wei Feng and Ivor Tsang and Song Wang},
  title     = {Leveraging Inpainting for Single-Image Shadow Removal},
  booktitle = {Proc. IEEE/CVF Int. Conf. Comput. Vis. (ICCV)},
  pages     = {13009--13018},
  year      = {2023},
  doi       = {10.1109/ICCV51070.2023.01200}
}

@inproceedings{liu2022convnet,
  author    = {Zhuang Liu and Hanzi Mao and Chao-Yuan Wu and Christoph Feichtenhofer and Trevor Darrell and Saining Xie},
  title     = {A {ConvNet} for the 2020s},
  booktitle = {Proc. IEEE/CVF Conf. Comput. Vis. Pattern Recog. (CVPR)},
  pages     = {11966--11976},
  year      = {2022},
  doi       = {10.1109/CVPR52688.2022.01167}
}

@inproceedings{m_Le-etal-ECCV18,
  author    = {Hieu Le and Tom{\'a}s F. Yago Vicente and Vu Nguyen and Minh Hoai and Dimitris Samaras},
  title     = {{A+D Net}: Training a Shadow Detector with Adversarial Shadow Attenuation},
  booktitle = {Proc. Eur. Conf. Comput. Vis. (ECCV)},
  pages     = {680--696},
  year      = {2018},
  doi       = {10.1007/978-3-030-01216-8_41}
}

@inproceedings{ntire2023wsrd,
  author    = {Florin-Alexandru Vasluianu and Tim Seizinger and Radu Timofte},
  title     = {{WSRD}: A Novel Benchmark for High Resolution Image Shadow Removal},
  booktitle = {Proc. IEEE/CVF Conf. Comput. Vis. Pattern Recog. Workshops (CVPRW)},
  pages     = {1826--1835},
  year      = {2023},
  doi       = {10.1109/CVPRW59228.2023.00181}
}

@inproceedings{okabe2009attached,
  author    = {Takahiro Okabe and Imari Sato and Yoichi Sato},
  title     = {Attached Shadow Coding: Estimating Surface Normals from Shadows under Unknown Reflectance and Lighting Conditions},
  booktitle = {Proc. IEEE Int. Conf. Comput. Vis. (ICCV)},
  pages     = {1693--1700},
  year      = {2009},
  doi       = {10.1109/ICCV.2009.5459381}
}

@inproceedings{panagopoulos2009robust,
  author    = {Alexandros Panagopoulos and Dimitris Samaras and Nikos Paragios},
  title     = {Robust Shadow and Illumination Estimation Using a Mixture Model},
  booktitle = {Proc. IEEE Conf. Comput. Vis. Pattern Recog. Workshops (CVPRW)},
  year      = {2009},
  doi       = {10.1109/CVPRW.2009.5206665}
}

@inproceedings{panagopoulos2011cvpr,
  author    = {Alexandros Panagopoulos and Chaohui Wang and Dimitris Samaras and Nikos Paragios},
  title     = {Illumination Estimation and Cast Shadow Detection through a Higher-Order Graphical Model},
  booktitle = {Proc. IEEE Conf. Comput. Vis. Pattern Recog. (CVPR)},
  pages     = {673--680},
  year      = {2011},
  doi       = {10.1109/CVPR.2011.5995585}
}

@inproceedings{panagopoulos2011iccv,
  author    = {Alexandros Panagopoulos and Tom{\'a}s F. Yago Vicente and Dimitris Samaras},
  title     = {Illumination Estimation from Shadow Borders},
  booktitle = {Proc. IEEE Int. Conf. Comput. Vis. Workshops (ICCVW)},
  pages     = {798--805},
  year      = {2011},
  doi       = {10.1109/ICCVW.2011.6130334}
}

@inproceedings{panagopoulos2012estimating,
  author    = {Alexandros Panagopoulos and Chaohui Wang and Dimitris Samaras and Nikos Paragios},
  title     = {Estimating Shadows with the Bright Channel Cue},
  booktitle = {Trends and Topics in Computer Vision},
  pages     = {1--12},
  year      = {2012},
  doi       = {10.1007/978-3-642-35740-4_1}
}

@article{panagopoulos2012simultaneous,
  author  = {Alexandros Panagopoulos and Chaohui Wang and Dimitris Samaras and Nikos Paragios},
  title   = {Simultaneous Cast Shadows, Illumination and Geometry Inference Using Hypergraphs},
  journal = {IEEE Trans. Pattern Anal. Mach. Intell.},
  volume  = {35},
  number  = {2},
  pages   = {437--449},
  year    = {2013},
  doi     = {10.1109/TPAMI.2012.110}
}

@inproceedings{paszke2019pytorch,
  author    = {Adam Paszke and Sam Gross and Francisco Massa and Adam Lerer and James Bradbury and Gregory Chanan and Trevor Killeen and Zeming Lin and Natalia Gimelshein and Luca Antiga and Alban Desmaison and Andreas K{\"o}pf and Edward Yang and Zach DeVito and Martin Raison and Alykhan Tejani and Sasank Chilamkurthy and Benoit Steiner and Lu Fang and Junjie Bai and Soumith Chintala},
  title     = {{PyTorch}: An Imperative Style, High-Performance Deep Learning Library},
  booktitle = {Adv. Neural Inf. Process. Syst. (NeurIPS)},
  volume    = {32},
  year      = {2019},
  doi       = {10.48550/arXiv.1912.01703},
  eprint    = {1912.01703},
  archivePrefix = {arXiv}
}

@inproceedings{patni2024ecodepth,
  author    = {Suraj Patni and Aradhye Agarwal and Chetan Arora},
  title     = {{ECoDepth}: Effective Conditioning of Diffusion Models for Monocular Depth Estimation},
  booktitle = {Proc. IEEE/CVF Conf. Comput. Vis. Pattern Recog. (CVPR)},
  pages     = {28285--28295},
  year      = {2024},
  doi       = {10.1109/CVPR52733.2024.02672}
}

@inproceedings{phongthawee2024diffusionlight,
  author    = {Pakkapon Phongthawee and Worameth Chinchuthakun and Nontaphat Sinsunthithet and Varun Jampani and Amit Raj and Pramook Khungurn and Supasorn Suwajanakorn},
  title     = {{DiffusionLight}: Light Probes for Free by Painting a Chrome Ball},
  booktitle = {Proc. IEEE/CVF Conf. Comput. Vis. Pattern Recog. (CVPR)},
  pages     = {98--108},
  year      = {2024},
  doi       = {10.1109/CVPR52733.2024.00018}
}

@inproceedings{piccinelli2024unidepth,
  author    = {Luigi Piccinelli and Yung-Hsu Yang and Christos Sakaridis and Mattia Segu and Siyuan Li and Luc Van Gool and Fisher Yu},
  title     = {{UniDepth}: Universal Monocular Metric Depth Estimation},
  booktitle = {Proc. IEEE/CVF Conf. Comput. Vis. Pattern Recog. (CVPR)},
  pages     = {10106--10116},
  year      = {2024},
  doi       = {10.1109/CVPR52733.2024.00963}
}

@inproceedings{Qu_2017_CVPR,
  author    = {Liangqiong Qu and Jiandong Tian and Shengfeng He and Yandong Tang and Rynson W. H. Lau},
  title     = {{DeshadowNet}: A Multi-Context Embedding Deep Network for Shadow Removal},
  booktitle = {Proc. IEEE Conf. Comput. Vis. Pattern Recog. (CVPR)},
  pages     = {2308--2316},
  year      = {2017},
  doi       = {10.1109/CVPR.2017.248}
}

@article{ranftl2020towards,
  author  = {Ren{\'e} Ranftl and Katrin Lasinger and David Hafner and Konrad Schindler and Vladlen Koltun},
  title   = {Towards Robust Monocular Depth Estimation: Mixing Datasets for Zero-Shot Cross-Dataset Transfer},
  journal = {IEEE Trans. Pattern Anal. Mach. Intell.},
  volume  = {44},
  number  = {3},
  pages   = {1623--1637},
  year    = {2022},
  doi     = {10.1109/TPAMI.2020.3019967}
}

@inproceedings{ranftl2021vision,
  author    = {Ren{\'e} Ranftl and Alexey Bochkovskiy and Vladlen Koltun},
  title     = {Vision Transformers for Dense Prediction},
  booktitle = {Proc. IEEE/CVF Int. Conf. Comput. Vis. (ICCV)},
  pages     = {12159--12168},
  year      = {2021},
  doi       = {10.1109/ICCV48922.2021.01196}
}

@article{sato2003illumination,
  author  = {Imari Sato and Yoichi Sato and Katsushi Ikeuchi},
  title   = {Illumination from Shadows},
  journal = {IEEE Trans. Pattern Anal. Mach. Intell.},
  volume  = {25},
  number  = {3},
  pages   = {290--300},
  year    = {2003},
  doi     = {10.1109/TPAMI.2003.1182093}
}

@incollection{Shadow_tracking,
  author    = {Pakorn KaewTraKulPong and Richard Bowden},
  title     = {An Improved Adaptive Background Mixture Model for Real-Time Tracking with Shadow Detection},
  booktitle = {Video-Based Surveillance Systems},
  pages     = {135--144},
  publisher = {Springer},
  year      = {2002},
  doi       = {10.1007/978-1-4615-0913-4_11}
}

@inproceedings{somanath2021hdr,
  author    = {Gowri Somanath and Daniel Kurz},
  title     = {{HDR} Environment Map Estimation for Real-Time Augmented Reality},
  booktitle = {Proc. IEEE/CVF Conf. Comput. Vis. Pattern Recog. (CVPR)},
  pages     = {11293--11301},
  year      = {2021},
  doi       = {10.1109/CVPR46437.2021.01114}
}

@inproceedings{sudre2017generalised,
  author    = {Carole H. Sudre and Wenqi Li and Tom Vercauteren and Sebastien Ourselin and M. Jorge Cardoso},
  title     = {Generalised Dice Overlap as a Deep Learning Loss Function for Highly Unbalanced Segmentations},
  booktitle = {Proc. Deep Learn. Med. Image Anal. Multimodal Learn. Clin. Decis. Support},
  pages     = {240--248},
  year      = {2017},
  doi       = {10.1007/978-3-319-67558-9_28}
}

@inproceedings{sun2023adaptive,
  author    = {Jiayu Sun and Ke Xu and Youwei Pang and Lihe Zhang and Huchuan Lu and Gerhard Hancke and Rynson Lau},
  title     = {Adaptive Illumination Mapping for Shadow Detection in Raw Images},
  booktitle = {Proc. IEEE/CVF Int. Conf. Comput. Vis. (ICCV)},
  pages     = {12663--12672},
  year      = {2023},
  doi       = {10.1109/ICCV51070.2023.01167}
}

@inproceedings{Sunkavalli2010VisibilitySU,
  author    = {Kalyan Sunkavalli and Todd Zickler and Hanspeter Pfister},
  title     = {Visibility Subspaces: Uncalibrated Photometric Stereo with Shadows},
  booktitle = {Proc. Eur. Conf. Comput. Vis. (ECCV)},
  pages     = {251--264},
  year      = {2010},
  doi       = {10.1007/978-3-642-15552-9_19}
}

@inproceedings{vasluianu2024ntire,
  author    = {Florin-Alexandru Vasluianu and Tim Seizinger and Zhuyun Zhou and Zongwei Wu and Cailian Chen and Radu Timofte and others},
  title     = {{NTIRE} 2024 Image Shadow Removal Challenge Report},
  booktitle = {Proc. IEEE/CVF Conf. Comput. Vis. Pattern Recog. Workshops (CVPRW)},
  pages     = {6547--6570},
  year      = {2024},
  doi       = {10.1109/CVPRW63382.2024.00654}
}

@inproceedings{verbin2024eclipse,
  author    = {Dor Verbin and Ben Mildenhall and Peter Hedman and Jonathan T. Barron and Todd Zickler and Pratul P. Srinivasan},
  title     = {{Eclipse}: Disambiguating Illumination and Materials Using Unintended Shadows},
  booktitle = {Proc. IEEE/CVF Conf. Comput. Vis. Pattern Recog. (CVPR)},
  pages     = {77--86},
  year      = {2024},
  doi       = {10.1109/CVPR52733.2024.00016}
}

@inproceedings{Vicente-etal-ECCV16,
  author    = {Tom{\'a}s F. Yago Vicente and Le Hou and Chen-Ping Yu and Minh Hoai and Dimitris Samaras},
  title     = {Large-Scale Training of Shadow Detectors with Noisily-Annotated Shadow Examples},
  booktitle = {Proc. Eur. Conf. Comput. Vis. (ECCV)},
  pages     = {816--832},
  year      = {2016},
  doi       = {10.1007/978-3-319-46466-4_49}
}

@inproceedings{Wang_2018_CVPR,
  author    = {Jifeng Wang and Xiang Li and Jian Yang},
  title     = {Stacked Conditional Generative Adversarial Networks for Jointly Learning Shadow Detection and Shadow Removal},
  booktitle = {Proc. IEEE/CVF Conf. Comput. Vis. Pattern Recog. (CVPR)},
  pages     = {1788--1797},
  year      = {2018},
  doi       = {10.1109/CVPR.2018.00192}
}

@inproceedings{wang2020instance,
  author    = {Tianyu Wang and Xiaowei Hu and Qiong Wang and Pheng-Ann Heng and Chi-Wing Fu},
  title     = {Instance Shadow Detection},
  booktitle = {Proc. IEEE/CVF Conf. Comput. Vis. Pattern Recog. (CVPR)},
  pages     = {1877--1886},
  year      = {2020},
  doi       = {10.1109/CVPR42600.2020.00195}
}

@inproceedings{Wang2020PeopleAS,
  author    = {Yifan Wang and Brian L. Curless and Steven M. Seitz},
  title     = {People as Scene Probes},
  booktitle = {Proc. Eur. Conf. Comput. Vis. (ECCV)},
  pages     = {438--454},
  year      = {2020},
  doi       = {10.1007/978-3-030-58607-2_26}
}

@inproceedings{wang2021single,
  author    = {Tianyu Wang and Xiaowei Hu and Chi-Wing Fu and Pheng-Ann Heng},
  title     = {Single-Stage Instance Shadow Detection with Bidirectional Relation Learning},
  booktitle = {Proc. IEEE/CVF Conf. Comput. Vis. Pattern Recog. (CVPR)},
  pages     = {1--11},
  year      = {2021},
  doi       = {10.1109/CVPR46437.2021.00007}
}

@article{wang2022instance,
  author  = {Tianyu Wang and Xiaowei Hu and Pheng-Ann Heng and Chi-Wing Fu},
  title   = {Instance Shadow Detection with a Single-Stage Detector},
  journal = {IEEE Trans. Pattern Anal. Mach. Intell.},
  volume  = {45},
  number  = {3},
  pages   = {3259--3273},
  year    = {2023},
  doi     = {10.1109/TPAMI.2022.3185628}
}

@article{wang2022pvt,
  author  = {Wenhai Wang and Enze Xie and Xiang Li and Deng-Ping Fan and Kaitao Song and Ding Liang and Tong Lu and Ping Luo and Ling Shao},
  title   = {{PVT} v2: Improved Baselines with Pyramid Vision Transformer},
  journal = {Comput. Vis. Media},
  volume  = {8},
  number  = {3},
  pages   = {415--424},
  year    = {2022},
  doi     = {10.1007/s41095-022-0274-8}
}

@article{wang2024swinshadow,
  author  = {Yonghui Wang and Shaokai Liu and Li Li and Wengang Zhou and Houqiang Li},
  title   = {{SwinShadow}: Shifted Window for Ambiguous Adjacent Shadow Detection},
  journal = {ACM Trans. Multimedia Comput. Commun. Appl.},
  volume  = {20},
  number  = {11},
  pages   = {1--20},
  year    = {2024},
  doi     = {10.1145/3688803}
}

@inproceedings{wang2025metashadow,
  author    = {Tianyu Wang and Jianming Zhang and Haitian Zheng and Zhihong Ding and Scott Cohen and Zhe Lin and Wei Xiong and Chi-Wing Fu and Luis Figueroa and Soo Ye Kim},
  title     = {{MetaShadow}: Object-Centered Shadow Detection, Removal, and Synthesis},
  booktitle = {Proc. IEEE/CVF Conf. Comput. Vis. Pattern Recog. (CVPR)},
  pages     = {28252--28262},
  year      = {2025},
  doi       = {10.1109/CVPR52734.2025.02631}
}

@inproceedings{Wu2025ImportanceBasedTM,
  author    = {Haoyu Wu and Jingyi Xu and Hieu Le and Dimitris Samaras},
  title     = {Importance-Based Token Merging for Efficient Image and Video Generation},
  booktitle = {Proc. IEEE/CVF Int. Conf. Comput. Vis. (ICCV)},
  pages     = {4983--4995},
  year      = {2025},
  doi       = {10.1109/ICCV51701.2025.00474}
}

@article{xing2024video,
  author  = {Zhenghao Xing and Tianyu Wang and Xiaowei Hu and Haoran Wu and Chi-Wing Fu and Pheng-Ann Heng},
  title   = {Video Instance Shadow Detection Under the Sun and Sky},
  journal = {IEEE Trans. Image Process.},
  volume  = {33},
  pages   = {5715--5726},
  year    = {2024},
  doi     = {10.1109/TIP.2024.3468877}
}

@inproceedings{Xu2022GeneratingRS,
  author    = {Jingyi Xu and Hieu Le},
  title     = {Generating Representative Samples for Few-Shot Classification},
  booktitle = {Proc. IEEE/CVF Conf. Comput. Vis. Pattern Recog. (CVPR)},
  pages     = {8993--9003},
  year      = {2022},
  doi       = {10.1109/CVPR52688.2022.00880}
}

@inproceedings{Xu2023FSOD,
  author    = {Jingyi Xu and Hieu Le and Dimitris Samaras},
  title     = {Generating Features with Increased Crop-Related Diversity for Few-Shot Object Detection},
  booktitle = {Proc. IEEE/CVF Conf. Comput. Vis. Pattern Recog. (CVPR)},
  pages     = {19713--19722},
  year      = {2023},
  doi       = {10.1109/CVPR52729.2023.01888}
}

@inproceedings{Xu2023ZeroShotOC,
  author    = {Jingyi Xu and Hieu Le and Vu Nguyen and Viresh Ranjan and Dimitris Samaras},
  title     = {Zero-Shot Object Counting},
  booktitle = {Proc. IEEE/CVF Conf. Comput. Vis. Pattern Recog. (CVPR)},
  pages     = {15548--15557},
  year      = {2023},
  doi       = {10.1109/CVPR52729.2023.01492}
}

@inproceedings{Xu2024ASQ,
  author    = {Jingyi Xu and Hieu Le and Dimitris Samaras},
  title     = {Assessing Sample Quality via the Latent Space of Generative Models},
  booktitle = {Proc. Eur. Conf. Comput. Vis. (ECCV)},
  pages     = {449--464},
  year      = {2024},
  doi       = {10.1007/978-3-031-73202-7_26}
}

@inproceedings{yang2023silt,
  author    = {Han Yang and Tianyu Wang and Xiaowei Hu and Chi-Wing Fu},
  title     = {{SILT}: Shadow-Aware Iterative Label Tuning for Learning to Detect Shadows from Noisy Labels},
  booktitle = {Proc. IEEE/CVF Int. Conf. Comput. Vis. (ICCV)},
  pages     = {12641--12652},
  year      = {2023},
  doi       = {10.1109/ICCV51070.2023.01165}
}

@inproceedings{yang2024depth,
  author        = {Lihe Yang and Bingyi Kang and Zilong Huang and Zhen Zhao and Xiaogang Xu and Jiashi Feng and Hengshuang Zhao},
  title         = {Depth Anything {V2}},
  booktitle     = {Adv. Neural Inf. Process. Syst. (NeurIPS)},
  volume        = {37},
  pages         = {21875--21911},
  year          = {2024},
  doi           = {10.52202/079017-0688},
  eprint        = {2406.09414},
  archivePrefix = {arXiv},
  primaryClass  = {cs.CV}
}

@inproceedings{yang2024depthanything,
  author    = {Lihe Yang and Bingyi Kang and Zilong Huang and Xiaogang Xu and Jiashi Feng and Hengshuang Zhao},
  title     = {Depth Anything: Unleashing the Power of Large-Scale Unlabeled Data},
  booktitle = {Proc. IEEE/CVF Conf. Comput. Vis. Pattern Recog. (CVPR)},
  pages     = {10371--10381},
  year      = {2024},
  doi       = {10.1109/CVPR52733.2024.00987}
}

@inproceedings{yang2025mli,
  author    = {Yixiong Yang and Shilin Hu and Haoyu Wu and Ramon Baldrich and Dimitris Samaras and Maria Vanrell},
  title     = {{mli-NeRF}: Multi-Light Intrinsic-Aware Neural Radiance Fields},
  booktitle = {Proc. Int. Conf. 3D Vision (3DV)},
  pages     = {587--596},
  year      = {2025},
  doi       = {10.1109/3DV66043.2025.00059}
}

@inproceedings{zhang2019all,
  author    = {Jinsong Zhang and Kalyan Sunkavalli and Yannick Hold-Geoffroy and Sunil Hadap and Jonathan Eisenman and Jean-Fran{\c{c}}ois Lalonde},
  title     = {All-Weather Deep Outdoor Lighting Estimation},
  booktitle = {Proc. IEEE/CVF Conf. Comput. Vis. Pattern Recog. (CVPR)},
  pages     = {10150--10158},
  year      = {2019},
  doi       = {10.1109/CVPR.2019.01040}
}

@article{Zhang2019ImprovingSS,
  author  = {Wuming Zhang and Xi Zhao and Jean-Marie Morvan and Liming Chen},
  title   = {Improving Shadow Suppression for Illumination Robust Face Recognition},
  journal = {IEEE Trans. Pattern Anal. Mach. Intell.},
  volume  = {41},
  number  = {3},
  pages   = {611--624},
  year    = {2019},
  doi     = {10.1109/TPAMI.2018.2803179}
}

@inproceedings{Zheng2019DistractionAwareSD,
  author    = {Quanlong Zheng and Xiaotian Qiao and Ying Cao and Rynson W. H. Lau},
  title     = {Distraction-Aware Shadow Detection},
  booktitle = {Proc. IEEE/CVF Conf. Comput. Vis. Pattern Recog. (CVPR)},
  pages     = {5162--5171},
  year      = {2019},
  doi       = {10.1109/CVPR.2019.00531}
}

@article{zheng2024bilateral,
  author  = {Peng Zheng and Dehong Gao and Deng-Ping Fan and Li Liu and Jorma Laaksonen and Wanli Ouyang and Nicu Sebe},
  title   = {Bilateral Reference for High-Resolution Dichotomous Image Segmentation},
  journal = {CAAI Artif. Intell. Res.},
  pages   = {9150038},
  year    = {2024},
  doi     = {10.26599/AIR.2024.9150038}
}

@inproceedings{zhu18b,
  author    = {Lei Zhu and Zijun Deng and Xiaowei Hu and Chi-Wing Fu and Xuemiao Xu and Jing Qin and Pheng-Ann Heng},
  title     = {Bidirectional Feature Pyramid Network with Recurrent Attention Residual Modules for Shadow Detection},
  booktitle = {Proc. Eur. Conf. Comput. Vis. (ECCV)},
  pages     = {122--137},
  year      = {2018},
  doi       = {10.1007/978-3-030-01231-1_8}
}

@article{Zhu2012ObjectbasedCA,
  author  = {Zhe Zhu and Curtis E. Woodcock},
  title   = {Object-Based Cloud and Cloud Shadow Detection in {Landsat} Imagery},
  journal = {Remote Sens. Environ.},
  volume  = {118},
  pages   = {83--94},
  year    = {2012},
  doi     = {10.1016/j.rse.2011.10.028}
}

@inproceedings{zhu2021mitigating,
  author    = {Lei Zhu and Ke Xu and Zhanghan Ke and Rynson W. H. Lau},
  title     = {Mitigating Intensity Bias in Shadow Detection via Feature Decomposition and Reweighting},
  booktitle = {Proc. IEEE/CVF Int. Conf. Comput. Vis. (ICCV)},
  pages     = {4682--4691},
  year      = {2021},
  doi       = {10.1109/ICCV48922.2021.00466}
}

@inproceedings{zhu2022single,
  author    = {Yurui Zhu and Xueyang Fu and Chengzhi Cao and Xi Wang and Qibin Sun and Zheng-Jun Zha},
  title     = {Single Image Shadow Detection via Complementary Mechanism},
  booktitle = {Proc. ACM Int. Conf. Multimedia (ACM MM)},
  pages     = {6717--6726},
  year      = {2022},
  doi       = {10.1145/3503161.3547904}
}
\end{document}